\documentclass[lettersize,journal]{IEEEtran}
\usepackage{amsmath,amsfonts}
\usepackage{algorithmic}
\usepackage{algorithm}
\usepackage{array}
\usepackage[caption=false]{subfig}
\usepackage{textcomp}
\usepackage{stfloats}
\usepackage{url}
\usepackage{verbatim}
\usepackage{graphicx}
\usepackage{cite}
\usepackage{amssymb}

\usepackage{array}
\usepackage{multirow}
\usepackage{diagbox}
\usepackage{tablefootnote}
\usepackage{soul}
\usepackage{pifont}
\usepackage{hyperref}       
\usepackage{amsmath,bm}
\usepackage{amsfonts}       
\usepackage{mathtools}
\usepackage{xcolor}

\usepackage{relsize}

\usepackage{etoolbox}
\makeatletter
\patchcmd{\@makecaption}
  {\scshape}
  {}
  {}
  {}
\makeatother

\begin{document}

\title{Efficient Heterogeneous Graph Learning via Random Projection}

\author{Jun Hu, Bryan Hooi, Bingsheng He
\IEEEcompsocitemizethanks{
\IEEEcompsocthanksitem Jun Hu, Bryan Hooi, and Bingsheng He are with the School of Computing, National University of Singapore, Singapore. (Corresponding author: Bryan Hooi)\protect\\
E-mail: jun.hu@nus.edu.sg; bhooi@comp.nus.edu.sg; hebs@comp.nus.edu.sg
}
}



\markboth{IEEE Transactions on Knowledge and Data Engineering}%
{Shell \MakeLowercase{\textit{et al.}}: A Sample Article Using IEEEtran.cls for IEEE Journals}


\maketitle

\begin{abstract}

Heterogeneous Graph Neural Networks (HGNNs) are powerful tools for deep learning on heterogeneous graphs.
Typical HGNNs require repetitive message passing during training, limiting efficiency for large-scale real-world graphs.
Recent pre-computation-based HGNNs use one-time message passing to transform a heterogeneous graph into regular-shaped tensors, enabling efficient mini-batch training.
Existing pre-computation-based HGNNs can be mainly categorized into two styles, which differ in how much information loss is allowed and efficiency.
We propose a hybrid pre-computation-based HGNN, named Random Projection Heterogeneous Graph Neural Network (RpHGNN), which combines the benefits of one style's efficiency with the low information loss of the other style.
To achieve efficiency, the main framework of RpHGNN consists of propagate-then-update iterations, where we introduce a Random Projection Squashing step to ensure that complexity increases only linearly.
To achieve low information loss, we introduce a Relation-wise Neighbor Collection component with an Even-odd Propagation Scheme, which aims to collect information from neighbors in a finer-grained way. 
Experimental results indicate that our approach achieves state-of-the-art results on seven small and large benchmark datasets while also being 230\% faster compared to the most effective baseline.  
Surprisingly, our approach not only surpasses pre-processing-based baselines but also outperforms end-to-end methods.
\footnote{Our code is available at: \url{https://github.com/CrawlScript/RpHGNN}.}

\end{abstract}


\section{Introduction}\label{sec:intro}

\IEEEPARstart{H}{eterogeneous} \textbf{graphs} are a class of graph structures that contain multiple types of vertices or edges. 
Heterogeneous graphs offer a richer and more informative representation compared to their homogeneous counterparts, where all vertices and edges are of the same type.
Heterogeneous graphs have gained popularity in recent years due to their ability to better represent complex relationships, with applications such as academic classification~\cite{9950622,DBLP:conf/kdd/HuDWCS20,bojchevski2020scaling,DBLP:journals/tkde/YuSDZL23} and social recommendation~\cite{9415142,DBLP:journals/tkde/LiWLS22,chen2023heterogeneous}.
In real-world scenarios, we frequently encounter large-scale heterogeneous graphs; however, dealing with them can become impractical due to out-of-memory (OOM) issues or excessively long processing times, e.g., spanning hours.
Thus, when developing techniques for handling heterogeneous graphs, the \textbf{efficiency} of the approach emerges as a critical consideration.

\begin{figure}[!tp]
\vspace{-4mm}
\centering\includegraphics[width=2.9in]{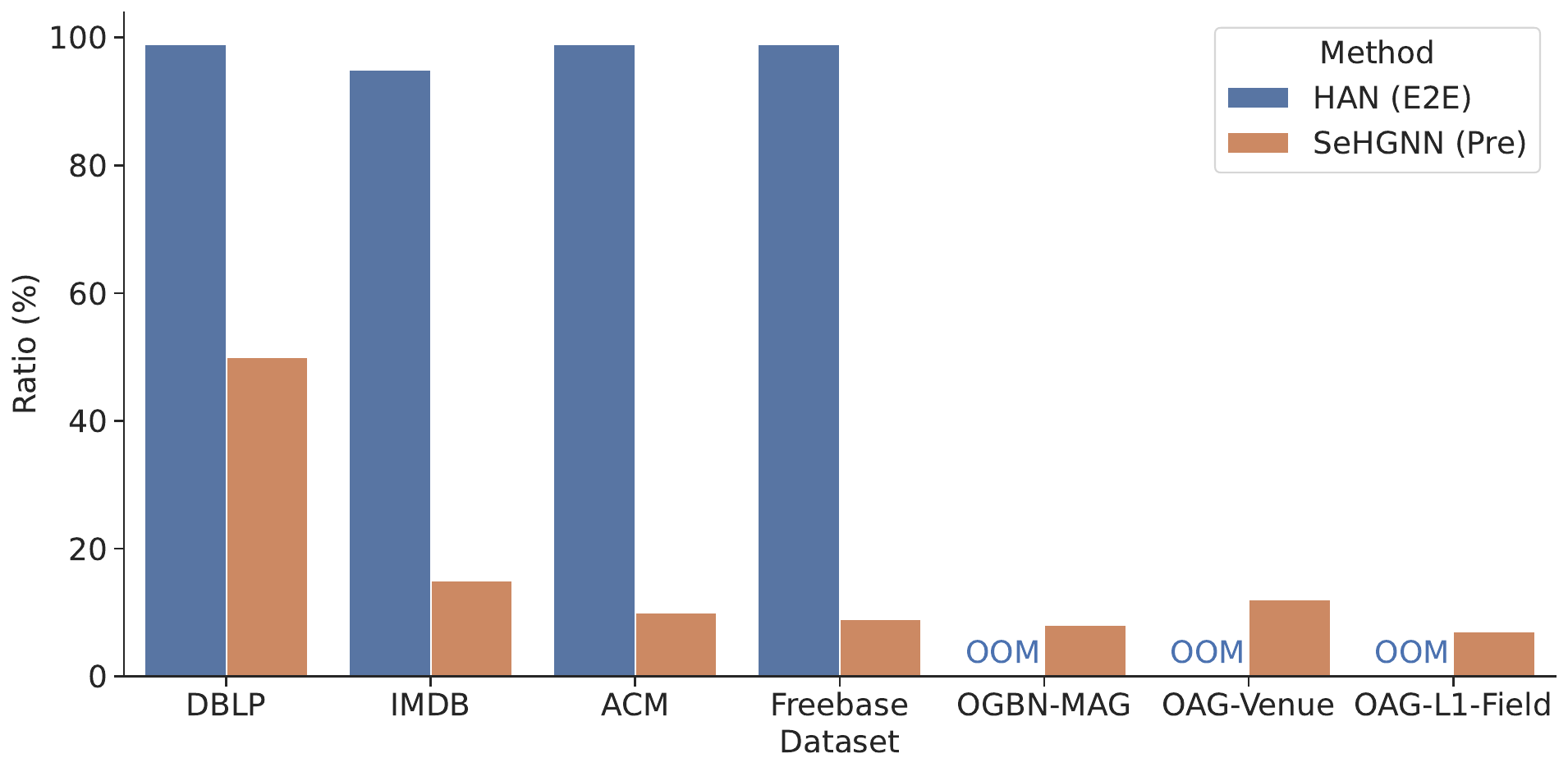}
\vspace{-4mm}
\caption{
Time consumption ratio of message passing to total for HAN (E2E) and SeHGNN (Pre).
(E2E) and (Pre) denote end-to-end and pre-computation-based HGNNs, respectively.
HAN (E2E) employs repetitive and resource-intensive message passing during training, constituting a remarkable 95\%-99\% of the total computation time.
Conversely, SeHGNN (Pre) utilizes a one-time message passing approach during pre-computation, leading to a markedly reduced ratio of message passing time relative to total computation time.
}
 \label{fig:message_passing_ratio}
\end{figure}

\begin{figure}[!tp]
\vspace{-2mm}
\centering\includegraphics[width=3.1in]{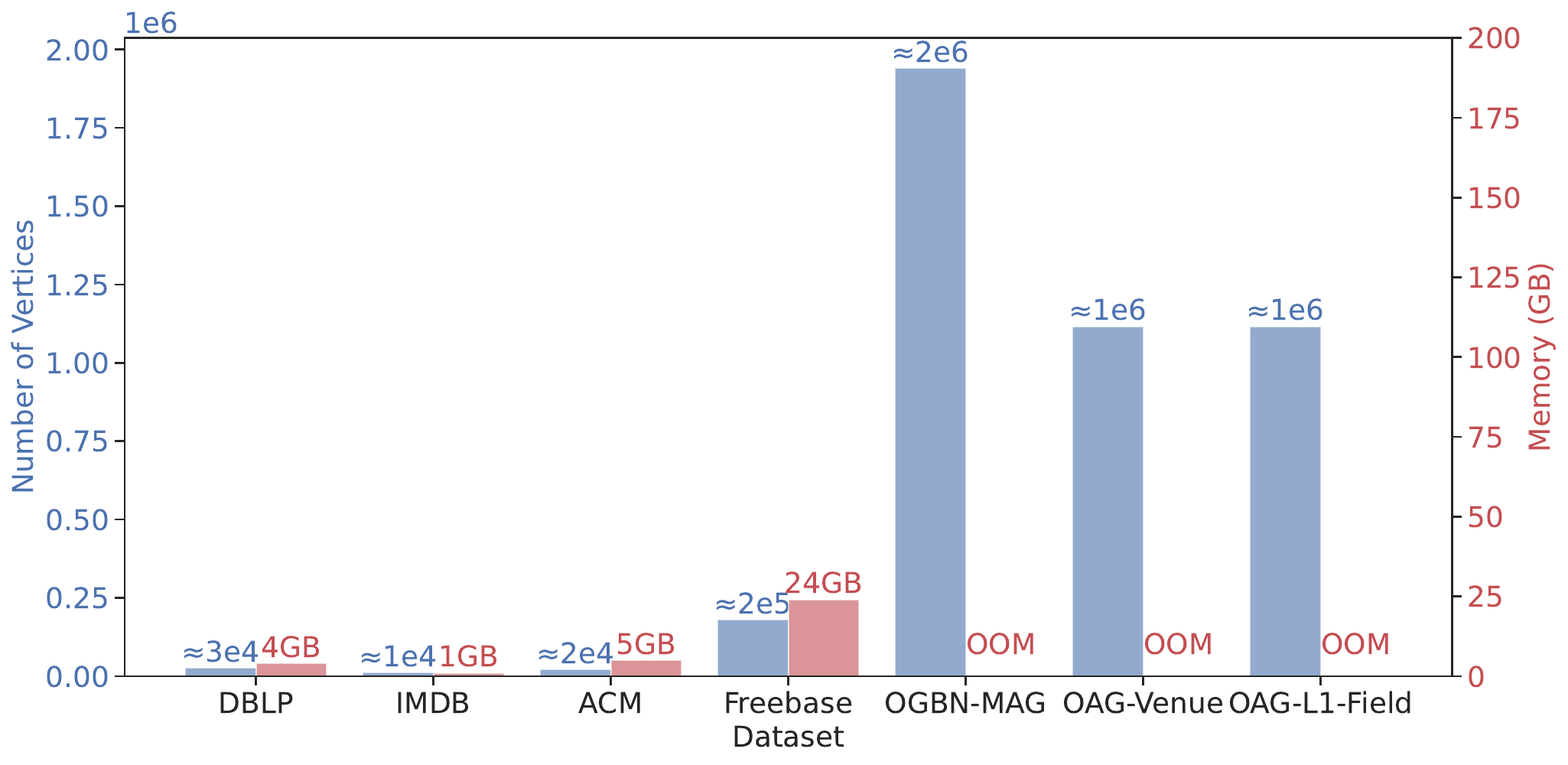}
\vspace{-4mm}
\caption{
Memory required by HAN for datasets of varying sizes (tested on a GPU with a 24GB memory capacity).
HAN is trained end-to-end and relies on the entire graph for message-passing, leading to costly computations for large-scale graphs.
As a result, HAN exhibits OOM errors on all large datasets (OGBN-MAG, OAG-Venue, and OAG-L1-Field).
Even when processing the significantly smaller dataset, Freebase, the GPU memory is fully consumed.
}
 \label{fig:memory}
 \vspace{-4mm}
\end{figure}

\textbf{Heterogeneous Graph Neural Networks (HGNNs)} are popular techniques for deep learning on heterogeneous graphs.
The main idea behind HGNNs is to aggregate valuable neighbor information based on a variety of relations to complement the semantics of vertices.
To achieve this, HGNNs typically rely on message passing, a process that gathers information (messages) from neighbors based on different relations (in the form of meta-paths, detailed in Section~\ref{sec:meta_path}) to enrich the semantic representations of vertices.
Typical HGNNs, like HAN~\cite{DBLP:conf/www/WangJSWYCY19}, are usually trained end-to-end (E2E).
They execute message passing during each training step (forward-backward step), leading to repetitive message passing throughout the training process.
As illustrated in Fig.~\ref{fig:message_passing_ratio}, the message passing operations in HAN (represented by the blue bars) account for a staggering 95\%-99\% of the total computation time.
Message passing has its limitations, because it depends on the entire graph for computation, which results in costly computations for large-scale graphs. 
Consequently, directly applying typical HGNNs to large-scale graphs is often impractical due to the \textbf{repetitive and resource-intensive message passing}.
Fig.~\ref{fig:memory} displays the memory required by HAN for datasets of varying sizes (only the number of vertices is shown for clarity), using a GPU with a maximum capacity of 24GB memory.
HAN exhibits OOM errors on all large datasets (OGBN-MAG, OAG-Venue, and OAG-L1-Field).
Even when processing the significantly smaller dataset, Freebase, the GPU memory is fully consumed.

\textbf{Pre-computation-based HGNNs}.
To overcome this limitation, various methods have been proposed in recent years, such as sampling-based~\cite{hamilton2017inductive,chen2018fastgcn,zeng2019graphsaint,huang2018adaptive} and pre-computation-based~\cite{yu2020scalable,yang2023simple} approaches. 
Among them, pre-computation-based approaches have shown state-of-the-art performance and the best overall efficiency on a variety of datasets.
Different from the repetitive message passing (on GPUs) during the training stage of typical HGNNs, pre-computation-based HGNNs design an efficient one-time message passing (on CPUs) to pre-process an entire heterogeneous graph into regular-shaped tensors for vertices. 
By using these regular-shaped tensors of vertices instead of the entire graph as inputs, HGNNs can easily leverage regular mini-batch techniques with simple encoders such as MLPs to implement efficient and scalable graph learning, where no further graph operations are required during training. 
The one-time message passing is typically executed on CPUs to conveniently handle large-scale graphs. 
Given that this message passing is only performed once during pre-computation, even on CPUs, pre-computation-based HGNNs exhibit a significantly lower ratio of message passing consumption time to total consumption time compared to typical end-to-end HGNNs, as illustrated in Fig.~\ref{fig:message_passing_ratio}.
In this paper, we focus on the pre-computation-based approach and present our novel pre-computation-based HGNN for efficient and scalable graph learning.

The \textbf{main challenge} of pre-computation-based HGNNs is how to \textbf{alleviate information loss} while striving to encapsulate as many details as possible from heterogeneous graphs into a limited number of regular-shaped tensors.
As with typical HGNNs, pre-computation-based HGNNs also incorporate the varying contributions of different relations (meta-paths) to a specific task.
This implies that the pre-computation stage should endeavor to retain details, such as neighbor information collected via different relations.
Preserving these details allows the training stage to balance the significance of varied relations, emphasizing important relations and suppressing irrelevant or noisy ones.
For the sake of efficiency, existing approaches~\cite{yu2020scalable} encapsulate the vast amount of details for the exponential number of relations into a limited number of regular-shaped tensors, which may cause severe \textbf{information loss} due to the untrainable nature of pre-computation.
(\textit{Untrainable nature}: operations during pre-computation, such as weighted vector sums, are untrainable.)

\begin{figure}[!tp]
\vspace{-4mm}
\centering
\subfloat[
Relation-wise Style.
This style collects neighbor information for all relations (meta-paths) within $K^{hop}$ hops separately, which is used to train HGNNs.\label{fig:relation_wise_style}
]{
\includegraphics[width=3.2in]{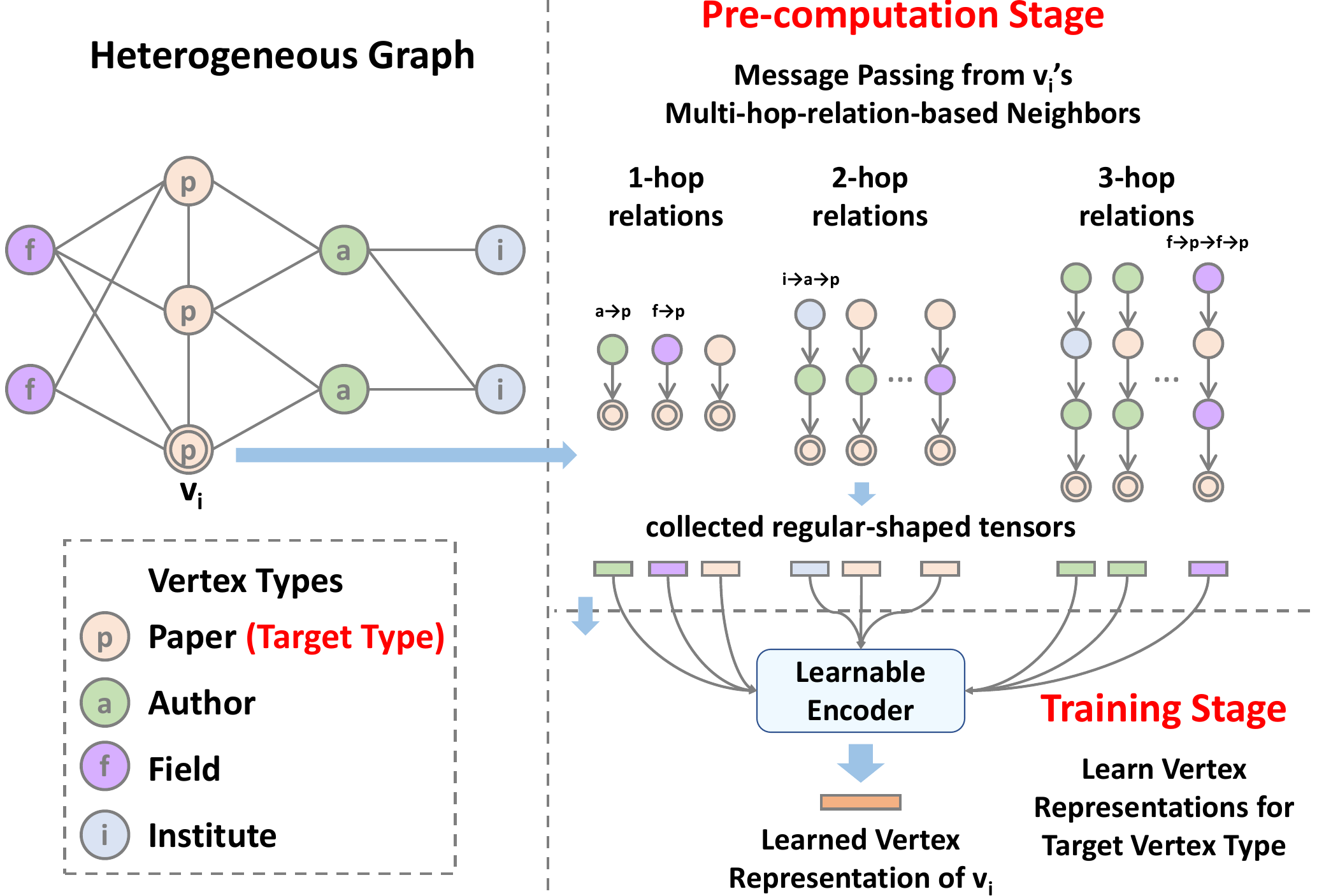}
}
\hfill
\subfloat[
Representation-wise Style.
This style consists of propagate-then-update iterations.
In each iteration, neighbor information is collected from local (1-hop) neighbors to update vertex representations.
The updated vertex representations accumulated across different iterations are used to train HGNNs.\label{fig:rep_wise_style}
]{
\includegraphics[width=3.2in]{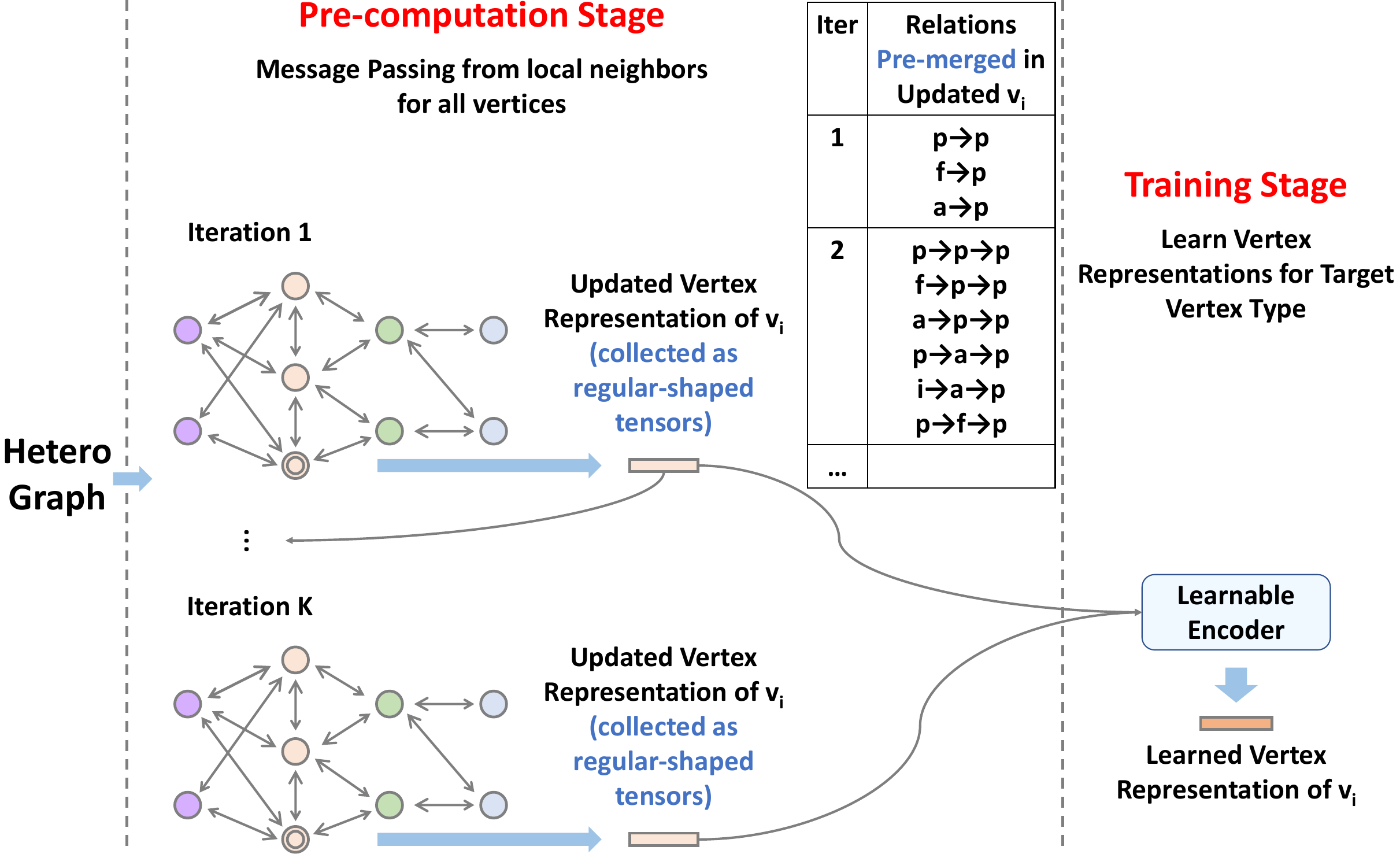}
}
\caption{Two main styles of pre-computation-based HGNNs.}
\label{fig:two_styles} 
\vspace{-6mm}
\end{figure}

\textbf{Two main styles} of pre-computation-based HGNNs.
Pre-computation-based HGNNS can be mainly categorized into two styles: relation-wise style and representation-wise style.
(1) \textbf{Relation-wise style}: As shown in Fig.~\ref{fig:relation_wise_style}, during pre-computation, this style handles different relations (in the form of meta-paths) separately and collects the neighbor information based on each relation into a dedicated vector.
Such fine-grained maintainence of various relations results in \textbf{low information loss}, and the subsequent learnable encoder can then flexibly learn to adjust the significance of these relations.
However, this advantage relies on the premise that the model can enumerate a sufficiently large number of relations, which typically increases exponentially with the number of hops $K^{hop}$. 
Therefore, this style may become \textbf{inefficient} when dealing with relations (both for pre-computation and learnable encoder) with $K^{hop} > 2$ on large-scale graphs.
(2) \textbf{Representation-wise style}: In contrast, the representation-wise style, as depicted in Fig.~\ref{fig:rep_wise_style}, employs several propagate-then-update iterations. 
Each iteration efficiently aggregates 1-hop neighbor messages to update vertex representations, which are collected as the input of the training stage. 
With $K$ iterations, the vertex representations can capture an exponential number of multi-hop relations within $K^{hop}=K$ hops, allowing the model to \textbf{efficiently} handle various relations with linear complexity.
However, as shown in the Table in Fig.~\ref{fig:rep_wise_style}, this style differs from the relation-wise style in that it may \textit{pre-merge} multiple relations into a single vector.
(\textit{Pre-merge}: for each vertex, the message passing in each iteration may involve neighbors based on different 1-hop relations, and all the messages are pre-merged into one single vector as the updated vertex representation, resulting in the pre-merge of relations.)
Such coarse-grained maintenance of relations hampers the subsequent learnable encoder's ability to adjust the relative importance of some pre-merged relations, leading to \textbf{high information loss}.

\begin{figure}[!tp]
\vspace{-2mm}
\centering\includegraphics[width=1.8in]{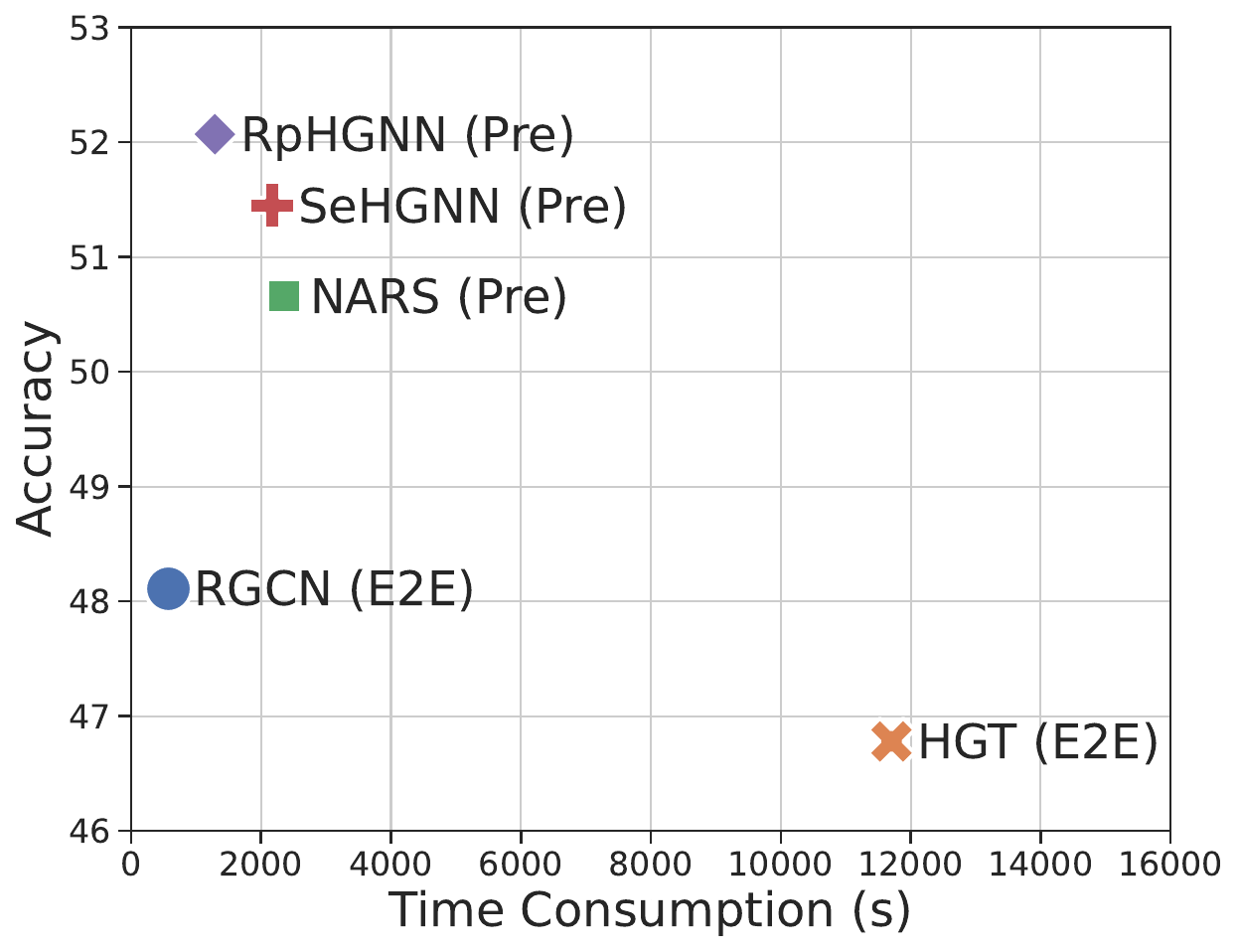}
\vspace{-2mm}
\caption{
Accuracy and time consumption of different HGNNs on a large dataset (OGBN-MAG).
(Pre) and (E2E) refer to pre-computation-based and end-to-end HGNNs, respectively.
}
 \label{fig:intro}
 \vspace{-4mm}
\end{figure}

\textbf{Our hybrid style pre-computation-based model.}
To exploit the advantage of both styles, in this paper, we propose a hybrid style pre-computation-based HGNN, Random Projection Heterogeneous Graph Neural Network (RpHGNN).
(1) To take advantage of \textbf{representation-wise style's efficiency}, RpHGNN's main framework adopts representation-wise style's propagate-then-update iterations, where we introduce a Random Projection Squashing step to ensure that the complexity increases only linearly.
The Random Projection Squashing step leverages random projection techniques to ensure that the dimensionality of the updated vertex representations remains constant across different iterations, thus maintaining efficiency.
(2) To take advantage of \textbf{relation-wise style's low information loss}, we introduce a Relation-Wise Neighbor Collection component with an Even-odd Propagation Scheme, which aims to collect information from
neighbors in a finer-grained way.
The even-odd propagation scheme enables our model to depend on fewer propagate-then-update iterations, consequently reducing the risk of information loss caused by untrainable vertex representation update operations within each iteration.
As shown in Fig.~\ref{fig:intro}, by effectively combining the two styles, our hybrid approach excels in overall performance and efficiency compared to existing approaches.

Our contributions are as follows:
\begin{itemize}
\item In this paper, we propose a hybrid style pre-computation-based HGNN, named Random Projection Heterogeneous Graph Neural Network (RpHGNN), which combines the benefits of representation-wise style's efficiency and relation-wise style's low information loss. 
\item To achieve \textbf{efficiency}, the main framework of RpHGNN consists of propagate-then-update iterations, where we introduce a Random Projection Squashing step to ensure that the complexity increases only linearly.
\item To achieve \textbf{low information loss}, we introduce a Relation-Wise Neighbor Collection component with an Even-odd Propagation Scheme, which aim to collect information from neighbors in a finer-grained way.
\item Experimental results indicate that our framework achieves state-of-the-art results on seven small and large benchmark datasets while also being 230\% faster compared to the most effective baseline (SeHGNN).  
Surprisingly, our approach outperforms not just the pre-computation-based baselines but also end-to-end baselines.
\end{itemize}

\section{Preliminaries}

\subsection{Heterogeneous Graphs.}

We use $G = (V, E, \mathcal{T}^{v}, \mathcal{T}^{e})$ to denote a heterogeneous graph, where $V$ and $E$ are the sets of vertices and edges, respectively, and $\mathcal{T}^{v}$ and $\mathcal{T}^{e}$ are the sets of vertex types and edge types, respectively.
$V=\bigcup\limits_{\tau^{v} \in \mathcal{T}^{v}} V^{\tau^{v}}$ may consist of different types of vertices, where $\tau^{v} \in \mathcal{T}^{v}$ is a vertex type, and $V^{\tau^{v}}$ denotes the set of vertices of vertex type $\tau^{v}$.
$E=\bigcup\limits_{\tau^{e} \in \mathcal{T}^{e}} E^{\tau^{e}}$ may consist of different types of edges, where $\tau^{e} \in \mathcal{T}^{e}$ denotes an edge type, and $E^{\tau^{e}}$ represents the set of edges of edge type $\tau^{e}$.
For vertex/edge types, each vertex type is a unique string, while each edge type is a unique tuple in the form $(\tau^{v}_i, \mathsf{unique\_name}, \tau^{v}_j)$, denoting this type of edges point from vertices of type $\tau^{v}_i$ to vertices of type $\tau^{v}_j$.
A unique name is also assigned to each edge type to help recognize it.
Given the edge type $\tau^{e} = (\tau^{v}_i, \mathsf{unique\_name}, \tau^{v}_j)$, we use $(\tau^{e})^{\prime} = (\tau^{v}_j, \mathsf{r.unique\_name}, \tau^{v}_i)$ to denote the reverse of $\tau^{e}$.

\textbf{Target vertex type / target vertex: } In many heterogeneous graph learning tasks, the objectives mainly only focus on a specific type of vertices.
For example, given an academic heterogeneous graph of vertex types $\mathtt{paper}$, $\mathtt{field}$, $\mathtt{author}$, and $\mathtt{institute}$, the objective of tasks such as paper classification may only care about the semantics/representations of $\mathtt{paper}$ vertices.
Therefore, we follow the settings of some benchmark datasets~\cite{lv2021we,hu2020open} and call the vertex type associated with the objective the target vertex type.
For brevity, we use the term \textbf{target vertex} to refer to a vertex of the target vertex type.

\subsection{Meta-Path and Meta-Path Schema}\label{sec:meta_path} 

\textbf{Meta-path Schema.}
A Meta-path is a path in a heterogeneous graph constrained by a Meta-path Schema.
We denote a meta-path schema of $L$ hops as follows:
\vspace{-1mm}
\begin{equation}
\vspace{-1mm}
    \mathfrak{mps}=\tau^{v}_{i_1} \xrightarrow[]{\tau^{e}_{j_1}} \tau^{v}_{i_2} \xrightarrow[]{\tau^{e}_{j_2}} ...\xrightarrow[]{\tau^{e}_{j_{L}}} \tau^{v}_{i_{(L+1)}}
\end{equation}
where $\tau^{v}_{i_l} \in \mathcal{T}^v$ is a node type which constrains the $l_{th}$ vertex of the path, and $\tau^{e}_{j_l} \in \mathcal{T}^e$ is an edge type which constrains the edge between the $l_{th}$ and $(l+1)_{th}$ vertices of the path.
For example, $\mathtt{paper} \xrightarrow[]{\mathsf{r.write}} \mathtt{author} \xrightarrow[]{\mathsf{write}} \mathtt{paper}$ is a meta-path schema.
It constrains the vertex types of the first, second, and third vertices in the meta-path to be $\mathtt{paper}$, $\mathtt{author}$, and $\mathtt{paper}$, respectively, and the edge types of the first and second edges to be $\mathsf{r.write}$ and $\mathsf{write}$, respectively.

\textbf{Meta-path.}
A path satisfying a given meta-path schema $\mathfrak{mps}$ is said to be a meta-path (instance) of $\mathfrak{mps}$.
Obviously, the number of meta-path schemas, as well as the number of meta-paths, grows exponentially as the number of hops $L$ increases, causing difficulty in designing scalable heterogeneous GNNs.

Meta-paths can capture meaningful \textbf{multi-hop relations} in heterogeneous graphs.
For example, the meta-paths of the meta-path schema $\mathtt{paper} \xrightarrow[]{\mathsf{r.write}} \mathtt{author} \xrightarrow[]{\mathsf{write}} \mathtt{paper}$ captures the co-author relations between papers.
Similarly, the meta-paths of the meta-path schema $\mathtt{paper} \xrightarrow[]{\mathsf{r.write}} \mathtt{author} \xrightarrow[]{\mathsf{write}} \mathtt{paper} \xrightarrow[]{\mathsf{has\_field}} \mathtt{field}$ can capture the research fields of co-authored papers.

\begin{figure}[!tp]
\vspace{-2mm}
\centering\includegraphics[width=1.8in]{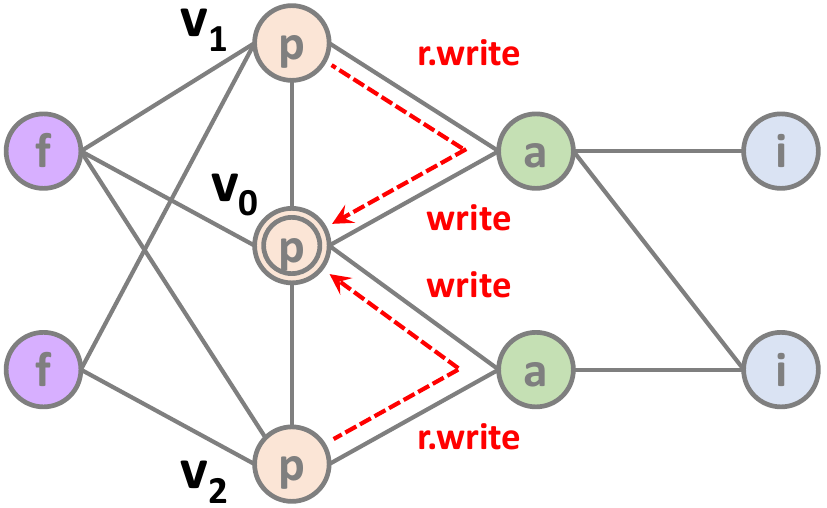}
\vspace{-2mm}
\caption{
An example of neighbors based on a relation (aka meta-path schema).
Given a $\mathtt{paper}$ vertex $v_0$, its neighbors based on the relation $\mathtt{paper} \xrightarrow[]{\mathsf{r.write}} \mathtt{author} \xrightarrow[]{\mathsf{write}} \mathtt{paper}$ are $v_1$ and $v_2$.
(The two red paths represent meta-paths constrained by the given relation.)
This indicates that both papers $v_1$ and $v_2$ are related to paper $v_0$ in terms of sharing an author with it.
}
 \label{fig:meta_path}
 \vspace{-4mm}
\end{figure}

\subsection{Relation}

\textbf{Relation.}
In the context of this paper, we adopt the term ``relation" as a shorthand for the more comprehensive term ``meta-path schema". 
To illustrate, when we discuss how relation-wise style HGNNs gather neighbor information for different relations separately, it denotes that they are gathering neighbor information for varied meta-path schemas separately.

A \textbf{local relation} is essentially a meta-path schema with a length or hop of 1, symbolically represented as $\tau^{v}_{i} \xrightarrow[]{\tau^{e}} \tau^{v}_{j}$.

\textbf{Neighbors based on a relation.}
In order to identify vertices that are related to a given vertex in a specific way (relation), we introduce the notion of neighbors based on a relation.
Formally, given a specific vertex and a specific relation (meta-path schema), we can identify a set of meta-paths that adhere to the given relation (meta-path schema) and end with the given vertex. 
We define the set of starting (left-most) vertices of these meta-paths as the neighbors based on the specific relation of the given vertex.
As depicted in Fig.~\ref{fig:meta_path}, for a given $\mathtt{paper}$ vertex $v_0$, co-authored $\mathtt{paper}$ vertices can be identified by finding neighbors of $v_0$ based on the relation $\mathtt{paper} \xrightarrow[]{\mathsf{r.write}} \mathtt{author} \xrightarrow[]{\mathsf{write}} \mathtt{paper}$.

\subsection{Neighbor Information Collection/Aggregation}
Given a specific vertex and relation, we can determine a set of neighbors based on the relation for the vertex.
We use the term ``collection/aggregation of neighbor information" to denote the operation that combines the feature vectors of these neighbors into a single vector, typically through operations such as mean pooling.

\section{Related Work}

\subsection{Graph Neural Networks}

Graph Neural Networks (GNNs) are specialized neural networks for graph deep learning.
GCN~\cite{kipf2016semi}, an early model, performs message passing in each layer to aggregate 1-hop neighbor information, enriching vertex semantics.
Stacking $K$ GCN layers allows GCNs to integrate information from neighbors within $K$ hops.
GAT~\cite{velivckovic2018graph} employs attention mechanisms to prioritize important neighbors during message passing.
GraphSAGE~\cite{hamilton2017inductive} enhances scalability for large graphs with neighbor sampling for mini-batch training. 
SGC~\cite{wu2019simplifying} simplifies multi-layer GNNs by removing non-linearities and weight matrices between layers, leading to faster processing without sacrificing accuracy.
APPNP~\cite{gasteiger2018predict} introduces an improved propagation scheme inspired by personalized PageRank~\cite{Page1999ThePC}, ensuring linear complexity via PageRank approximation.
S$^2$GC~\cite{DBLP:conf/iclr/ZhuK21} uses a modified Markov Diffusion Kernel for propagation, balancing global and local vertex contexts.

However, these GNNs show limitations in many real-world graphs due to challenges like heterophily~\cite{DBLP:conf/icml/Pan023,li2024pc,zhu2021graph} and heterogeneity.
In this paper, we focus on the challenge of heterogeneity, specifically that the aforementioned GNNs are tailored for homogeneous graphs and overlook the characteristics of heterogeneous graphs, where vertices and edges can be of different types and contribute differently.
HGNNs have been developed to address this by accommodating the properties of heterogeneous graphs.

\subsection{Heterogeneous Graph Neural Networks}

HGNNs can be roughly classified into two categories: relation-wise style and representation-wise style.

Relation-wise style HGNNs focus on identifying neighbors of target vertices based on different relations (meta-path schemas) separately and then aggregate this information to learn the target vertices representations. 
HAN~\cite{DBLP:conf/www/WangJSWYCY19} employs hand-designed meta-path schemas to distinguish between different neighbor semantics and applies a semantic attention mechanism to enhance GAT for neighbor aggregation. 
MAGNN~\cite{DBLP:conf/www/0004ZMK20} involves all vertices within a meta-path for enhanced semantics. 
HetGNN~\cite{zhang2019heterogeneous} samples neighbors via random walk and aggregates same-type vertex neighbors through Bi-LSTMs, behaving like a relation-wise style HGNN.

Representation-wise style HGNNs, on the other hand, involve several propagate-then-update iterations to iteratively perform message passing (between 1-hop neighbors) and vertex representation update (for all vertices). 
R-GCN~\cite{schlichtkrull2018modeling} extends GCN by aggregating neighbors based on distinct 1-hop relations using different transformation matrices for each relation. 
RSHN~\cite{zhu2019relation} designs a coarsened line graph to capture global embeddings of different edge types for neighbor aggregation. 
HetSANN~~\cite{hong2020attention} extends GAT by using type-specific score functions to generate attention values for different relations. 
HGT~\cite{DBLP:conf/www/HuDWS20} applies a heterogeneous mutual attention mechanism with type-specific parameters. 
Simple-HGN~\cite{lv2021we} extends GAT by including vertex features and learnable edge-type embeddings for attention generation.
HINormer~\cite{DBLP:conf/www/MaoLLS23} leverages a Graph Transformer with a larger-range aggregation mechanism to effectively capture both structural and heterogeneous information.

However, these HGNNs are end-to-end models and can be impractical for large-scale graphs due to repetitive and resource-intensive message passing during training.

\subsection{Pre-computation-based Heterogeneous Graph Neural Networks}

In addressing the efficiency challenges of HGNNs on large graphs, several strategies have emerged, such as sampling~\cite{ying2018graph,DBLP:conf/kdd/JiangJFSLW21} and pre-computation-based methods.
This paper particularly focuses on pre-computation-based methods, as they have demonstrated leading performance and optimal efficiency across various datasets.

We have introduced the two main styles of pre-computation-based HGNNs in Section~\ref{sec:intro}: relation-wise and representation-wise.
SeHGNN~\cite{yang2023simple}, as shown in Fig.~\ref{fig:relation_wise_style}, exemplifies the relation-wise style, collecting neighbor information of target vertices based on different relations within $K^{hop}$ hops separately, achieving low information loss but compromising efficiency.
As the number of relations increases exponentially with $K^{hop}$, SeHGNN can become inefficient when handling relations (both for pre-computation and the learnable encoder), especially when $K^{hop} > 2$ on large-scale graphs.
For instance, in the OAG-Venue dataset, the numbers of relations (considering reversed edge types) are 28, 272, and 2136 for $K^{hop}=1, 2$, and $3$, respectively.
Particularly, when $K^{hop}=3$, SeHGNN has to collect and maintain 2136 vectors per vertex, becoming impractical due to the OOM problem.
Fig.~\ref{fig:rep_wise_style} shows a basic representation-wise style pre-computation-based HGNN, SIGN~\cite{frasca2020sign}, which is efficient but may incur high information loss due to the pre-merge of relations as discussed in Section~\ref{sec:intro}.
(SIGN is designed for homogeneous graphs but can also be applied to heterogeneous graphs.)
To mitigate the high information loss problem, NARS~\cite{yu2020scalable} proposes to extract multiple subgraphs based on different sampled relation subsets. 
Instead of pre-computation on the entire heterogeneous graph as in Fig.~\ref{fig:rep_wise_style}, NARS does so on each sampled subgraph.

This paper proposes a hybrid pre-computation-based HGNN that seeks to combine the low information loss of the relation-wise style with the efficiency of the representation-wise style.

\section{Method}

\subsection{Overall Framework}

\begin{figure*}[!t]
\vspace{-2mm}
\centering\includegraphics[width=5.4in]{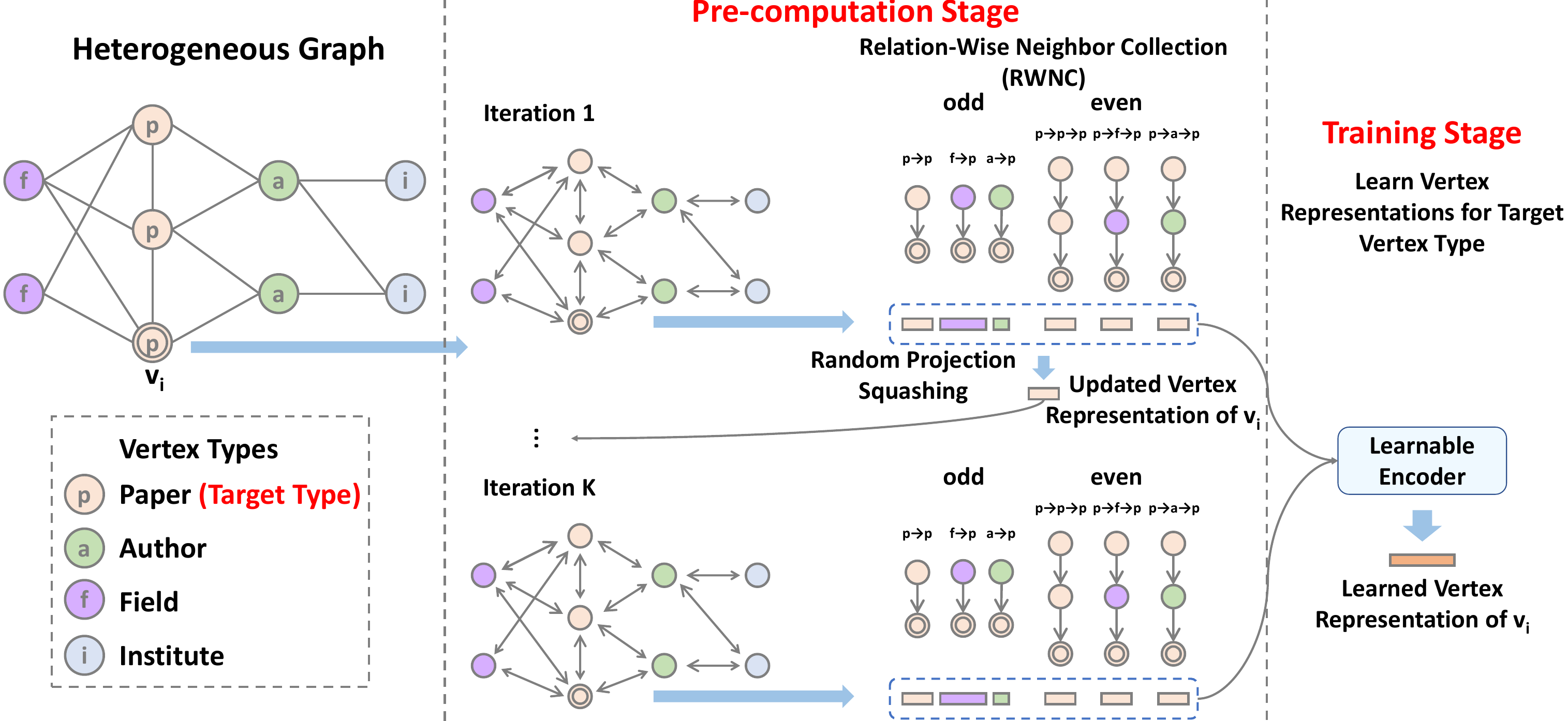}
\vspace{-1mm}
\caption{
Overall Framework of the proposed hybrid pre-computation-based HGNN, Random Projection Heterogeneous Graph Neural Network (RpHGNN).
(1) Pre-computation stage:
To leverage the representation-wise style’s efficiency, the main framework of RpHGNN consists of propagate-then-update iterations, where we introduce a Random Projection Squashing component to reduce dimensionality for updated vertex representations, ensuring linear complexity growth.
To exploit the relation-wise style’s fine granularity for low information loss, we introduce a Relation-wise Neighbor Collection component with an Even-odd Propagation Scheme, which collects neighbor information based on different local (or low-hop) relations separately.
(2) Training Stage:
After $K$ iterations, the collected neighbor information (of target vertices) is used as the input for a learnable encoder to learn the vertex representations of target vertices.
}
\label{fig:framework}
 \vspace{-2mm}
\end{figure*}

Like other pre-computation-based HGNNs, our framework has a pre-computation stage and a training stage (see Fig.~\ref{fig:framework}).
RpHGNN mainly consists of the following components:

\textit{Pre-computation Stage -} \textbf{Hybrid Propagate-then-update Iterations (HPI):}
To take advantage of the representation-wise style's \textbf{efficiency}, RpHGNN's main framework adopts the representation-wise style's propagate-then-update iterations, ensuring that our framework's complexity only increases linearly.
Each hybrid propagate-then-update iteration consists of:
\begin{itemize}
\item The \textbf{Relation-Wise Neighbor Collection (RWNC)} component collects neighbor information based on different local (or low-hop) relations separately, which can take advantage of the relation-wise style's fine granularity for \textbf{low information loss}.
An \textbf{even-odd propagation} scheme is further proposed to replace the local propagation scheme for low information loss.
\item The \textbf{Random Projection Squashing} component squashes the various neighbor information collected above for untrainable vertex representations updates.
It employs random projection for dimensionality reduction, thus avoiding excessive growth in dimensionality and ensuring \textbf{efficiency}.
\end{itemize}

\textit{Training Stage -} \textbf{Learnable Encoder:}
For target vertices, their various neighbor information, which is collected by the RWNC component over $K$ iterations, is used by the training stage to learn the vertex representations.
The vertex representations are learned via a simple learnable encoder consisting of Conv1D and fully-connected layers.

\subsection{Relation-Wise Neighbor Collection (RWNC)}

In each propagate-then-update iteration, RWNC collects neighbor information via message passing (propagation) to achieve two objectives:
\begin{itemize}
\item For target vertices, the collected neighbor information is used by the training stage to encode vertex representations.
\item For all vertices, the collected neighbor information is used for the untrainable vertex representation update in the current propagate-then-update iteration.
\end{itemize}

To take advantage of the relation-wise style's fine granularity for low information loss, we design RWNC to collect neighbor information based on different local (or low-hop) relations separately.
Note that in the final version, as illustrated in Fig.~\ref{fig:framework}, RWNC adopts an even-odd propagation scheme, which considers not only local relations but also some specific 2-hop relations (termed even relations).
For brevity, we introduce RWNC under the setting of a local propagation scheme.
We define the \textbf{``local propagation scheme"} as a protocol that limits message passing to collecting information solely from neighbors based on local relations.

Formally, given a vertex type $\tau^{v}_{j}$, we can identify a set of edge types ending with it, denoted as $\{\tau^{e}_{m} | \tau^{e}_{m} = (*, *, \tau^{v}_{j})\}$.
Each edge type $\tau^{e}_{m} = (\tau^{v}_{i}, *, \tau^{v}_{j})$ can be treated as a local relation $\tau^{v}_{i} \xrightarrow[]{\tau^{e}_{m}} \tau^{v}_{j}$, and our model performs message passing along it to collect neighbor information as follows:
\vspace{-1mm}
\begin{equation}\label{eq:local_prop}
\vspace{-1mm}
    (H^{\tau^{e}_{m}})^{(k)} = (D^{\tau^{e}_{m}})^{-1}A^{\tau^{e}_{m}} (H^{\tau^{v}_{i}})^{(k-1)}
\end{equation}
where, $(H^{\tau^{v}_{i}})^{(k-1)} \in \mathbb{R}^{|V^{\tau^{v}_{i}}| \times d^{\tau^{v}_{i}}}$ denotes the vertex representation matrix produced by the $(k-1)_{th}$ iteration (if $k > 1$) or the raw feature matrix (if $k = 1$) of vertices of type $\tau^{v}_{i}$.
Here, $|V^{\tau^{v}_{i}}|$ represents the number of vertices of type $\tau^{v}_{i}$, and $d^{\tau^{v}_{i}}$ denotes the dimensionality of the raw feature vectors (as well as the representation vectors) for these vertices.
$A^{\tau^{e}_{m}} \in \mathbb{R}^{|V^{\tau^{v}_{j}}| \times |V^{\tau^{v}_{i}}|}$ is the adjacency matrix corresponding to the edge type $\tau^{e}_{m}$.
Each entry, $A^{\tau^{e}_{m}}_{ab}$, within this matrix is either 1 or 0, indicating the presence or absence, respectively, of an edge that originates from the $b_{th}$ vertex of type $\tau^{v}_{i}$ and targets the $a_{th}$ vertex of type $\tau^{v}_{j}$.
$D^{\tau^{e}_{m}} \in \mathbb{R}^{|V^{\tau^{v}_{j}}| \times |V^{\tau^{v}_{j}}|}$ is the degree matrix for vertices of type $\tau^{v}_{j}$ with respect to the relation $\tau^{e}_{m}$.
It is a diagonal matrix and $D^{\tau^{e}_{m}}_{aa} = \sum \limits_{b=1}^{|V^{\tau^{v}_{i}}|} A^{\tau^{e}_{m}}_{ab}$.

We perform the above neighbor information collection for each edge type (local relation) $\tau^{e}_{m} \in \{\tau^{e}_{m} | \tau^{e}_{m} = (*, *, \tau^{v}_{j})\}$ separately.
For example, as depicted in the RWNC part of Fig.~\ref{fig:framework} (currently we should ignore the relations in the even part, which are introduced by the even-odd propagation scheme), given the vertex type $\tau^{v}_{j}=\mathtt{paper}$, we collect neighbor information for relations $\mathtt{paper} \xrightarrow[]{\mathsf{cite}} \mathtt{paper}$, $\mathtt{author} \xrightarrow[]{\mathsf{write}} \mathtt{paper}$, and $\mathtt{field} \xrightarrow[]{\mathsf{r.has\_field}} \mathtt{paper}$, separately.

In terms of the even-odd propagation scheme, given that its motivation is grounded in the understanding of the untrainable vertex representation update component (Random Projection Squashing), we will introduce it in Section~\ref{sec:even_odd_prop} after the introduction of Random Projection Squashing in Section~\ref{sec:rand_proj_squash}.

\subsection{Random Projection Squashing}\label{sec:rand_proj_squash} 

After RWNC collects various neighbor information based on different relations for each vertex, we need to squash them into one vector as the updated vertex representation.
The vertex representation update is non-trivial since the various collected neighbor information may have different dimensionality.
Naive solutions such as the concatenation of various neighbor information may cause the dimensionality of vertex representations to increase exponentially, resulting in severe efficiency issues.
To handle this challenge, we design a \textbf{Random Projection Squashing} technique, which utilizes random projection techniques for dimensionality reduction. 
This ensures that the dimensionality of the updated vertex representations remains constant, thus maintaining efficiency.

Here, we choose random projection as the dimensionality reduction technique due to its simplicity and computational efficiency. 
In our approach, a data matrix of size $N \times d_0$ is reduced to $N \times d_1$, where $N$ (with $N \gg d_0$) typically represents the number of vertices and $d_0 > d_1$. 
Random projection involves only weight matrix randomization and matrix multiplication, offering extreme simplicity with a computational complexity of $O(Nd_0d_1)$. 
The sparse random projection implementation~\cite{achlioptas2001database} uses a binary (0,1) weight matrix instead of a floating-point weight matrix, enabling the use of integer arithmetic to further reduce complexity. 
Other popular dimensionality reduction techniques, such as PCA~\cite{jolliffe2016principal}, may require complex eigen-decomposition, potentially increasing complexity to $O(Nd_0^2)$ and making them less efficient than random projection.
Additionally, since the data pre-processing steps of some baselines (e.g., NARS~\cite{yu2020scalable}) may already utilize random projection, we adopt it for consistency and to maintain simplicity in our overall experimental implementation.

For simplicity, we introduce our Random Projection Squashing component with the local propagation scheme instead of the even-odd propagation scheme. 
Specifically, given a vertex type $\tau^{v}_{j}$ and its collected neighbor information $\{(H^{\tau^{e}_{m}})^{(k)} | \tau^{e}_{m} = (*, *, \tau^{v}_{j})\}$, we update the vertex representation of $\tau^{v}_{j}$ as follows:
\vspace{-1mm}
\begin{equation}\label{eq:random_project_squash}
\vspace{-1mm}
\begin{aligned}
    (H^{\tau^{v}_{j}})^{(k)} = \sum \limits_{\tau^{e}_{m} = (*, *, \tau^{v}_{j})} \bm{\mathrm{Norm}}((H^{\tau^{e}_{m}})^{(k)} (W^{\tau^{e}_{m}})^{(k)})
\end{aligned}
\end{equation}
where $(W^{\tau^{e}_{m}})^{(k)} \in \mathbb{R}^{d_{*} \times d^{\tau^{v}_{j}}}$ is the \textbf{random projection weight matrix}, and $d^{\tau^{v}_{j}}$ is the dimensionality of the raw feature vectors of vertices of type $\tau^{v}_{j}$.
By default, we adopt an efficient sparse random projection implementation~\cite{achlioptas2001database} to generate $(W^{\tau^{e}_{m}})^{(k)}$ as follows:
\vspace{-2mm}
\begin{equation}\label{eq:rand_proj_squash}
\vspace{-1mm}
(W^{\tau^{e}_{m}})^{(k)}_{ab} = \begin{cases}
    1 & \text{with probability } \frac{1-p^{sp}}{2} \\
    0 & \text{with probability } p^{sp} \\
    -1 & \text{with probability } \frac{1-p^{sp}}{2}  \\
\end{cases}
\end{equation}
where $p^{sp}$ is a hyperparameter deciding the sparsity of the random projection weight matrix.
In Equation~\ref{eq:rand_proj_squash}, $\bm{\mathrm{Norm}}$ is the row-wise normalization operation, implemented as L2 normalization, normalizing each row $\mathbf{x}$ of a matrix as follows:
\vspace{-2mm}
\begin{equation}
\vspace{-2mm}
    \bm{\mathrm{Norm}}(\mathbf{x}) = \mathbf{x} / \lVert \mathbf{x} \rVert_2
\end{equation}
The random projection and the sum operation (sum pooling) in Equation~\ref{eq:random_project_squash} ensure that the dimensionality of the updated vertex representation \textbf{remains constant ($d^{\tau^{v}_{j}}$)}, thus maintaining efficiency.
In contrast, the naive solution for vertex representation update, which simply concatenates neighbor information collected via different relations, suffers from exponentially growing dimensionality in vertex representations.

Normalization operations mitigate the issue of \textbf{information overwhelming} when squashing neighbor information collected through various relations. 
Neighbor information can have varying dimensionalities. 
Although we employ random projection techniques to equalize the dimensionalities, they do not ensure stable output magnitude. 
As a result, input feature vectors with higher dimensionality might overwhelm those with lower dimensionality during sum pooling due to their excessive magnitude. 
To tackle this, we employ normalization to stabilize the magnitude of random projection outputs, thereby alleviating the information overwhelming issue.

Untrainable vertex representation updates may cause \textbf{information loss}.
During the untrainable pre-computation stage, data (tensors) may encounter information loss after applying operations such as tensor addition, tensor normalization, and tensor multiplication.
As with representation-wise style HGNNs, untrainable vertex representation updates involve the pre-merging of relations, which can result in information loss due to tensor addition and normalization operations.
Furthermore, random projection operations perform untrainable transformations with randomized weights, which may also lead to information loss due to tensor multiplication operations.
To alleviate the impact of such information loss caused by untrainable vertex representation updates, we propose the even-odd propagation scheme, which is introduced in the next section.

Note that random projection is also used in other heterogeneous graph learning research for efficiency improvements, such as FAME~\cite{DBLP:conf/cikm/LiuHYFD20} and NARS~\cite{yu2020scalable}.
These methods typically apply random projection operations to input vertex feature vectors just once, converting them into low-dimensional feature vectors before all other operations to improve efficiency for subsequent operations that rely on vertex features.
Unlike these methods, which utilize random projection to control input dimensionalities, our random projection operations are used to prevent excessive growth of dimensionality during propagate-then-update iterations of pre-computation.
Specifically, RpHGNN continually applies random projection operations to intermediate data with increased dimensionality produced by RWNC during propagate-then-update iterations, effectively preventing excessive growth in dimensionality and ensuring that the dimensionality of the updated vertex representations remains constant.

\subsection{Even-Odd Propagation Scheme}\label{sec:even_odd_prop}

\textbf{Goal: reduce the frequency} of untrainable vertex representation update operations, as these may lead to information loss.
The untrainable vertex representation update operation may cause information loss due to the pre-merge of relations and random projection operations.
This operation is executed at the end of each propagate-then-update iteration; thus, using more iterations increases the frequency of this operation, thereby increasing the risk of information loss.
We propose an even-odd propagation scheme, which allows us to use fewer propagate-then-update iterations without compromising the number of relation hops we can exploit.
Consequently, it reduces the frequency of untrainable vertex representation update operations, alleviating the risk of information loss.

Extend the local propagation scheme for fewer propagate-then-update iterations:
With the local propagation scheme, each iteration collects neighbor information via 1-hop relations, and thus $K$ propagate-then-update iterations are required to capture relations within $K$ hops.
A naive solution to reduce the number of iterations is to use a $2$-hop propagation scheme, which extends local relations to \textbf{relations within $2$ hops}.
This requires only $\frac{K}{2}$ iterations for relations within $K$ hops, reducing the frequency of untrainable vertex representation updates.
However, this solution may encounter \textbf{an efficiency problem}, since it is not uncommon that there are an overwhelming number of relations within 2 hops on certain graphs.

\textbf{Even/odd relations:}
To address the above efficiency problem, we design an even-odd propagation scheme, which only selects twice as many relations within 2 hops as local ones, including odd relations and even relations.
Specifically, given a vertex type  $\tau^{v}_j$ and an edge type ending with it $\tau^{e} = (\tau^{v}_i, *, \tau^{v}_j)$, we define the corresponding odd and even relations as follows:
\begin{itemize}
\item \textbf{Odd relation} is equivalent to the 1-hop (local) relation $\tau^{v}_{i} \xrightarrow[]{\tau^{e}} \tau^{v}_{j}$.
\item \textbf{Even relation} is defined as $\tau^{v}_{j} \xrightarrow[]{(\tau^{e})^{\prime}} \tau^{v}_{i} \xrightarrow[]{\tau^{e}} \tau^{v}_{j}$, which is a symmetric 2-hop meta-path since $(\tau^{e})^{\prime}$ is the reverse of $\tau^{e}$.
\end{itemize}

Neighbor information collection of odd/even relations:
Under the setting of the even-odd propagation scheme, we collect neighbor information for the odd relation and even relation of each edge type, respectively.
Given an odd relation $\tau^{v}_{i} \xrightarrow[]{\tau^{e}} \tau^{v}_{j}$, we collect the corresponding neighbor information $(H^{\tau^{e}_{m}})^{(k, odd)}$ in the same way as for $(H^{\tau^{e}_{m}})^{(k)}$ in Equation~\ref{eq:local_prop}.
Given an even relation $\tau^{v}_{j} \xrightarrow[]{(\tau^{e})^{\prime}} \tau^{v}_{i} \xrightarrow[]{\tau^{e}} \tau^{v}_{j}$, we can also collect the corresponding neighbor information via efficient sparse-dense matrix multiplication as follows:
%
%
\begin{equation}
    (H^{\tau^{e}_{m}})^{(k, even)} = (D^{\tau^{e}_{m}})^{-1}A^{\tau^{e}_{m}} (D^{(\tau^{e}_{m})^{\prime}})^{-1}A^{(\tau^{e}_{m})^{\prime}} 
 (H^{\tau^{v}_{j}})^{(k-1)}
\end{equation}

\vspace{-2mm}
\subsection{Learnable Encoder}\label{sec:learnable_encoder}

During the training stage, an encoder is trained to learn vertex representations using the collected neighbor information as input. 
Since the neighbor information is collected into regular-shaped tensors, the learnable encoder can benefit from \textbf{mini-batching} for efficient training.

\begin{figure}[!tp]
\vspace{-2mm}
\centering\includegraphics[width=3.1in]{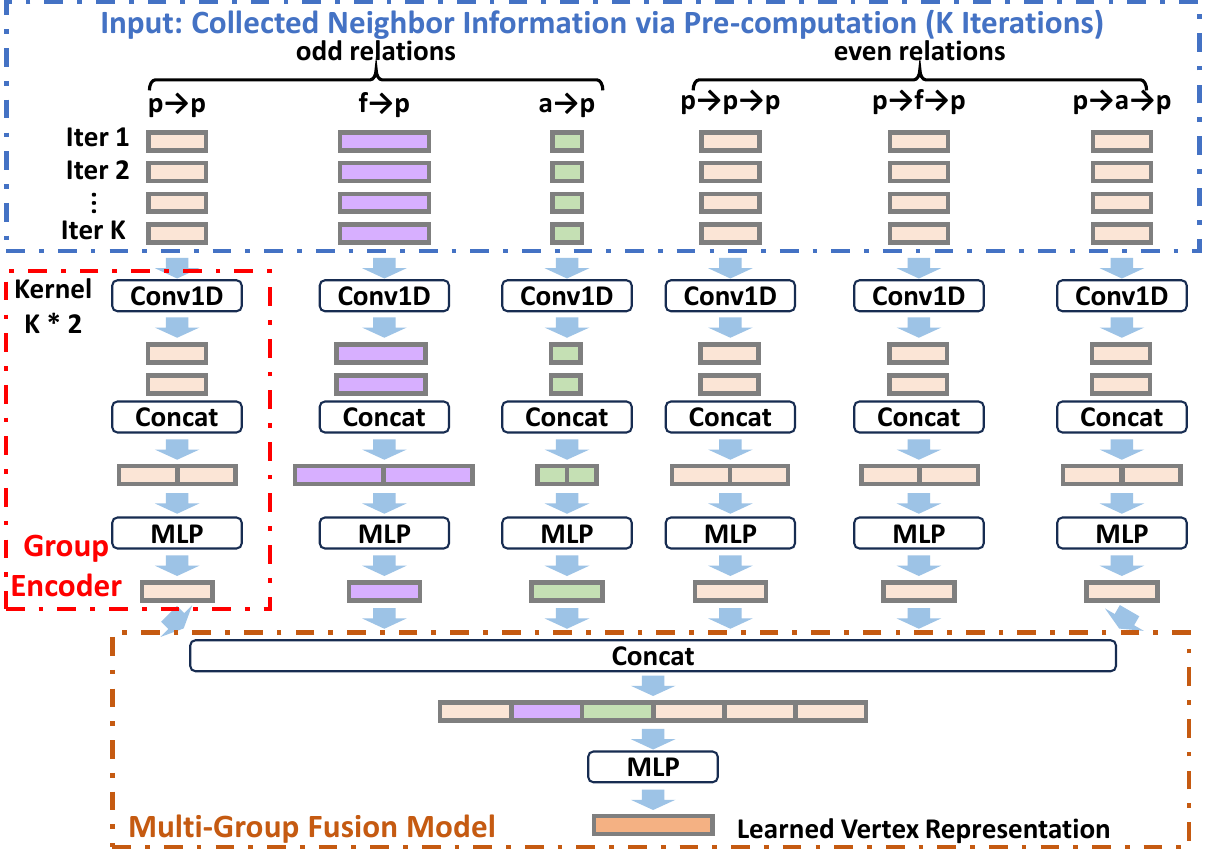}
\vspace{-2mm}
\caption{
The learnable encoder \textbf{takes as input the neighbor inforamtion collected via pre-computaion}, where neighbor information collected by the same relation is put into the same group (column).
In each group, we stack together neighbor information collected by different iterations (as rows).
The learnable encoder primarily consists of two components: 
(1) The \textbf{Group Encoder} is designed to encode neighbor information inside each group (relation) into a group representation. 
The group encoders adopt independent parameters across different groups, aiming to extract group-specific (relation-specific) information. 
(2)  The \textbf{Multi-group Fusion Model} takes the multiple encoded group representations as input and outputs the final vertex representation, aiming at exploiting interactions between different groups (relations) for vertex representation learning.
}
 \label{fig:learnable_encoder}
 \vspace{-4mm}
\end{figure}

Input of the learnable encoder:
For each target vertex, each propagate-then-update iteration collects its neighbor information based on several even and odd relations, and there are multiple propagate-then-update iterations.
The upper part of Fig.~\ref{fig:learnable_encoder} shows an example of how we organize this information.
We put neighbor information collected by the \textbf{same relation} into the \textbf{same group (column)}.
In each group, we stack neighbor information collected by different iterations together, where the $k_{th}$ row corresponds to the neighbor information collected by the $k_{th}$ iteration.

The main framework of our learnable encoder primarily consists of two components:
\begin{itemize}
\item The \textbf{Group Encoder} is designed to encode neighbor information inside each group (relation) into a group representation.
The group encoders adopt independent parameters across different groups, aiming to extract group-specific (relation-specific) information. 
Each group encoder consists of a 1D-Convolutional (Conv1D) layer, a concatenation operation, and a MLP model.
The Conv1D layer aims to integrate neighbor information collected by different iterations, and thus it considers the iteration indices as input channels.
We use $C^{out}$ to denote the number of output channels of the Conv1D operations, where the Conv1D is as follows:
\vspace{-2mm}
\begin{equation}
\vspace{-2mm}
M_{c}^{\tau^{v}_{j}} = \sum \limits_{k=1}^{K} W^{Conv}_{k,c} (H^{\tau^{v}_{j}})^{(k)} + \mathbf{b}_{c}
\end{equation}
where $W^{Conv} \in \mathbb{R}^{K \times C^{out}}$ and $\mathbf{b} \in \mathbb{R}^{C^{out}}$ are the weights and the bias of the Conv1D layer, respectively. 
The Conv1D layer outputs $C^{out}$ different vector representations $\{M_{c}^{\tau^{v}_{j}} | 1 \le c \le C^{out}\}$, aiming to provide different weights assigned to different iterations.
We then concatenate these different vector representations and encode them via an MLP to obtain the group representation.
\item The \textbf{Multi-Group Fusion Model} takes the multiple encoded group representations as input and outputs the final vertex representation.
Obviously, the multi-group fusion model can exploit interactions between different groups (relations) for vertex representation learning.
The implementation simply concatenates the multiple encoded group representations into a single representation vector, and then encodes it into the final vertex representation via an MLP model. 
\end{itemize}

\section{Analysis}

In this section, we perform a detailed analysis of the even-odd propagation scheme and the training stage complexity.

\subsection{Analysis of Even-Odd Propagation Scheme}

In this section, we begin by exploring the significance of even relations within heterogeneous graphs, followed by an illustrative example that highlights the benefits of employing our Even-Odd Propagation Scheme.

\subsubsection{The Significance of Even Relations in Heterogeneous Graphs}\label{sec:sig_of_even}
In heterogeneous graphs, neighbor information collected based on even relations may exhibit significant importance.
Considering a vertex of type $\tau^{v}_j$ and an edge type ending with it $\tau^{e} = (\tau^{v}_i, *, \tau^{v}_j)$, it's neighbors captured by the odd relation $\tau^{v}_{i} \xrightarrow[]{\tau^{e}} \tau^{v}_{j}$ and neighbors captured by the even relation $\tau^{v}_{j} \xrightarrow[]{(\tau^{e})^{\prime}} \tau^{v}_{i} \xrightarrow[]{\tau^{e}} \tau^{v}_{j}$ differ not only structurally, but also in terms of \textbf{the quality of the associated features}.
To illustrate this, let us consider a vertex of type $\mathtt{paper}$ and an edge of type $(\mathtt{author}, \mathsf{write}, \mathtt{paper})$.
\begin{itemize}
\item The neighbors based on the odd relation $\mathtt{author} \xrightarrow[]{\mathsf{write}} \mathtt{paper}$ are $\mathtt{author}$ vertices. 
Given the scarcity of author attribute data, we resort to using a random embedding vector as the feature vector for each author vertex. 
It is evident that this type of feature is of relatively low quality.
\item On the other hand, the neighbors based on the even relation $\mathtt{paper} \xrightarrow[]{\mathsf{r.write}} \mathtt{author} \xrightarrow[]{\mathsf{write}} \mathtt{paper}$ are $\mathtt{paper}$ vertices. 
For each $\mathtt{paper}$ vertex, we use the Doc2Vec~\cite{le2014distributed} representation of its title as the feature vector. 
Compared to the author vertices, the doc2vec features of the $\mathtt{paper}$ vertices demonstrate a significantly superior quality.
\end{itemize}
Consequently, the neighbor information aggregated via even relations might demonstrate greater importance compared to that collected through odd relations, suggesting that even relations could be vital in offering a more comprehensive understanding of heterogeneous graphs.

\subsubsection{Even-Odd Propagation Scheme Affects Information Loss}\label{sec:assign_pre_merge}
We compare the local and even-odd propagation schemes comprehensively via examples in Appendix~\ref{app:even-odd}, which show that the even-odd propagation scheme can reduce information loss by adopting fewer untrainable vertex representation updates.

\begin{table}[!tp]
\vspace{-2mm}
\centering
\caption{
Theoretical complexity of NARS, SeHGNN, and RpHGNN in every  training epoch.
$N$ is the number of target vertices.
For simplicity, we use the same $D$ as the dimensionality of input features and hidden representations of MLPs.
For NARS, $S$ is the number of sampled relation subsets.
For SeHGNN, $M$ is the number of relations (meta-path schemas) used.
For RpHGNN, $R$ is the number of relations that end with the target vertex type.
}
\scalebox{0.75}{
\begin{tabular}{|l |c| c| c|}
\hline
       & Group Encoder   & Multi-Group Fusion & Total  \\\hline
NARS   & $O(NKSD+NKD^2)$ & $O(NKD^2)$         & $O(NKSD+NKD^2)$ \\\hline
SeHGNN & $O(NMD^2)$      & $O(NM^2D^2)$       & $O(NM^2D^2)$  \\\hline
RpHGNN & $O(NKRD+NRD^2)$ & $O(NRD^2)$         & $O(NKRD+NRD^2)$ \\\hline
\end{tabular}
}
\label{tab:analysis_complexity}
\vspace{-2mm}
\end{table}

\subsection{Complexity Analysis of The Training Stage}

Table~\ref{tab:training_time} reveals that in large datasets (OGBN-MAG, OAG-Venue, OAG-L1-Field), pre-computation-based methods (NARS, SeHGNN, RpHGNN) primarily consume time during the training stage, impacting overall efficiency.

NARS, SeHGNN, and RpHGNN have comparable approaches in organizing input of the training stages. 
RpHGNN, as discussed in Section~\ref{sec:learnable_encoder}, organizes the neighbor information collected by the pre-computation into different groups of vectors. 
With $R$ relations ending in the target vertex type and $K$ propagate-then-update iterations, RpHGNN collects $2R$ groups, each containing $K$ vectors for each vertex.
NARS and SeHGNN adopt similar ways, with slight differences:
\begin{itemize}
\item \textbf{NARS} collects information from $K$ hops of neighbors across $S$ subgraphs, yielding $K$ groups, each consisting of $S$ vectors.
\item \textbf{SeHGNN} separately collects information based on the $M$ relations within $K$ hops, resulting in $M$ groups, each containing one vector for per vertex.
\end{itemize}

NARS and SeHGNN employ similar main frameworks for the learnable encoders as RpHGNN.
Their learnable encoders first adopt multiple group encoders to encode each group separately, and then adopt  a multi-group fusion model captures inter-group interactions to derive the final vertex representations.
These encoders generally use MLPs and 1D convolutional layers. 
However, SeHGNN's multi-group fusion model utilizes a Transformer model (citation needed), which may contribute to higher complexity.
Table~\ref{tab:analysis_complexity} lists the theoretical complexities of NARS, SeHGNN, and RpHGNN per training epoch, indicating that NARS and RpHGNN share comparable complexity levels.

\section{Experiments}

\subsection{Dataset and Task}

We evaluate our approach on several heterogeneous graph datasets, including DBLP, IMDB, ACM, Freebase, OGBN-MAG, OAG-Venue, and OAG-L1-Field.
Among them, DBLP, IMDB, ACM, and Freebase are small heterogeneous graph datasets used by the HGB benchmark~\cite{lv2021we}; OGBN-MAG, OAG-Venue, and OAG-L1-Field are large heterogeneous graph datasets used by popular scalable HGNN baselines~\cite{yu2020scalable,yang2023simple}.
The statistics of the datasets are listed in Table~\ref{tab:datasets_statistics}.

\textbf{Vertex classification tasks.}
These datasets are designed for vertex classification tasks.
Each dataset contains a target vertex type, and our task is to learn to predict the labels of vertices of this target vertex type.
All the datasets provide fixed data splitting (into training, validation, and test sets) for vertex classification tasks.

\subsection{Baselines}

We compare RpHGNN with both end-to-end (E2E) HGNN baselines and pre-computation-based (Pre) HGNN baselines.

The \textbf{end-to-end HGNN baselines} include several homogeneous GNNs: GCN~\cite{kipf2016semi}, GAT~\cite{velivckovic2018graph}, and GraphSAGE~\cite{hamilton2017inductive}; and several heterogeneous GNNs: RGCN~\cite{schlichtkrull2018modeling}, HAN~\cite{DBLP:conf/www/WangJSWYCY19}, GTN~\cite{yun2019graph}, RSHN~\cite{zhu2019relation}, HetGNN~\cite{zhang2019heterogeneous}, MAGNN~\cite{DBLP:conf/www/0004ZMK20}, HetSANN~\cite{hong2020attention}, HGT~\cite{DBLP:conf/www/HuDWS20}, Simple-HGN~\cite{lv2021we}, and HINormer~\cite{DBLP:conf/www/MaoLLS23}.
Note that most of them encounter out-of-memory (OOM) issues when dealing with large datasets.
Three exceptions are GraphSAGE, HGT and RGCN.
Both GraphSAGE and HGT propose their own neighbor sampling strategies, which allow them to handle large datasets.
As for RGCN, while the original version presents OOM issues with large datasets, it is often used as a baseline for large-scale HGNNs~\cite{yu2020scalable}.
Therefore, in an effort to adapt RGCN for large datasets, we follow NARS' experimental setting~\cite{yu2020scalable} and adopt the neighbor sampling strategy used by GraphSAGE.

The \textbf{pre-computation-based HGNN baselines} include NARS~\cite{yu2020scalable} and SeHGNN~\cite{yang2023simple}, which are representation-wise style and relation-wise style pre-computation-based HGNNs, respectively.
Note that in the original paper, SeHGNN refers to the full version that leverages a label propagation technique to exploit label information.
For clarity and consistency in our study, we modify the nomenclature from the original paper, using SeHGNN to refer to the version without label propagation and SeHGNN-LP for the version that incorporates it.
Note that label propagation is a general technique~\cite{xiaojin2008semi,zhou2003learning,huang2020combining} that is compatible with most GNNs and is not the focus of this paper.
Therefore, it is not used in our main experiments and is considered only in Section~\ref{sec:leaderboard}, where we demonstrate the successful integration of our model with other techniques.

\begin{table}[!tp]
\vspace{-2mm}
\centering
\caption{Statistics of Datasets.}
\vspace{-2mm}
\scalebox{0.75}{
\small
\begin{tabular}{|>{\centering\arraybackslash}m{1.9cm}|c| >{\centering\arraybackslash}m{7mm}|c|>{\centering\arraybackslash}m{7mm}|>{\centering\arraybackslash}m{1cm}|>{\centering\arraybackslash}m{1cm}|}\hline
Dataset      & Vertices   & Vertex Types & Edges      & Edge Types & Target Type       & Target Classes   \\\hline
DBLP         & 26,128     &      4       & 239,566    & 3          & $\mathtt{author}$ & 4   \\\hline
IMDB         & 21,420     &      4       & 86,642     & 3          & $\mathtt{movie}$  & 5   \\\hline
ACM          & 10,942     &      4       & 547,872    & 4          & $\mathtt{paper}$  & 3   \\\hline
Freebase     & 180,098    &      8       & 1,057,688  & 36         & $\mathtt{book}$   & 7   \\\hline
OGBN-MAG     & 1,939,743  &      4       & 21,111,007 & 4          & $\mathtt{paper}$ & 349 \\\hline
OAG-Venue    & 1,116,162  &      5       & 13,985,692 & 15         & $\mathtt{paper}$ & 3505 \\\hline
OAG-L1-Field & 1,116,163  &      5       & 13,016,549 & 15         & $\mathtt{paper}$ & 275  \\\hline
\end{tabular}
}
\label{tab:datasets_statistics}
\vspace{-3mm}
\end{table}

\subsection{Parameter Settings and Evaluation Metrics}

\subsubsection{Handling of Featureless Vertices}\label{sec:param_setting_random_emb}
In GNNs, the handling of featureless vertices affects both performance and efficiency. 
For end-to-end baselines, we follow the settings recommended by the HGB benchmark~\cite{lv2021we} and leverage learnable embeddings as input features for featureless vertices. 
However, in the case of pre-computation-based HGNNs, the propagation of vertex features occurs during an untrainable pre-computation stage, which \textbf{prevents us from utilizing learnable embeddings} as vertex features.

To address this, several strategies are available. 
One option is to use pre-trained embeddings learned through traditional Network Representation Learning (NRL) methods such as TransE~\cite{bordes2013translating,yu2020scalable} or ComplEx~\cite{DBLP:conf/icml/TrouillonWRGB16,yang2023simple}. 
Another approach is to employ Gaussian random embeddings~\cite{yang2023simple}. 
The former strategy, while effective, requires an additional time-consuming stage for pre-training the vertex embeddings. 
For instance, the pre-training of TransE on OGBN-MAG can take up to 40 minutes~\cite{yu2020scalable}, which may exceed the total time of our proposed approach. 
Considering these factors, we opt for the use of random embeddings for both pre-computation-based baselines and our method.

One exception is the end-to-end HGT method on large datasets, which relies on a sampling strategy to enable large-scale training. 
The official implementation of the sampling-based HGT (an end-to-end baseline) involves a complex multi-process feature sampling strategy, making it difficult to adopt learnable embeddings for featureless vertices. 
Therefore, similar to pre-computation-based approaches, we choose to adopt the random embedding strategy for HGT on large datasets.

\subsubsection{Training Settings}

For training, we follow the settings of the HGB benchmark~\cite{lv2021we} and utilize the performance improvement on the validation set as a beacon to determine whether the model has been improved.
Specifically, if a performance improvement on the validation set is observed during training, we replace the final model parameters with the current ones.
In addition, we also adopt the early stopping strategy used by the HGB benchmark: if no performance improvement on the validation set is observed within a specific count (referred to as patience), the training will be stopped early before reaching the maximum epochs.
This strategy assists in preventing overfitting, while also contributing to the computational efficiency of our model training.
We train our model 10 times with different random seeds and report the mean performance and time consumption.

There are some hyperparameters for training, including the learning rate $lr$ and batch size $bs$.
Since our model can already perform well with a nearly unified training hyperparameter setting, we do not tune these hyperparameters.
The unified setting is: $lr = 3e-3$, $bs=3000$ (on OAG-Venue and OAG-L1-Field datasets), and $bs=10000$ (on other datasets).
The reason we adopt a different batch size on OAG-Venue and OAG-L1-Field is that setting $bs=10000$ on these datasets may cause an OOM issue.

In terms of the running environment, we train the models on a Linux machine with 2 Intel(R) Xeon(R) CPU E5-2690 v4 @ 2.60GHz, 128GB RAM, and a GeForce GTX 1080 Ti with 11GB of GPU memory.

\subsubsection{Parameter Settings for Baselines}
In the case of RGCN, HAN, GTN, RSHN, HetGNN, MAGNN, HetSANN, HGT, GCN, GAT, Simple-HGN, and SeHGNN, the literature ~\cite{lv2021we,yang2023simple} provides tuned parameter settings along with corresponding performance on small datasets, namely DBLP, IMDB, ACM, and Freebase.
We, therefore, utilize these tuned parameter settings for these models on small datasets and adopt the reported performance.

In contrast, for certain scenarios where official results are not available (for instance, NARS on small datasets), or where the experimental settings differ (such as alternative strategies for handling featureless vertices, as mentioned in Section~\ref{sec:param_setting_random_emb}), we employ the official or benchmark implementations
\footnote{\url{https://github.com/THUDM/HGB}}
\footnote{\url{https://github.com/Ffffffffire/HINormer}}
\footnote{\url{https://github.com/dmlc/dgl/tree/master/examples/pytorch/ogb/ogbn-mag}}
\footnote{\url{https://github.com/UCLA-DM/pyHGT}}
\footnote{\url{https://github.com/facebookresearch/NARS}}
\footnote{\url{https://github.com/ICT-GIMLab/SeHGNN}} 
of the baseline models and diligently tune their hyperparameters to the best of our abilities.

\subsubsection{Parameter Settings for Our Approach}

In our approach, model hyperparameters are tuned based on the performance exhibited on the validation set. 
We conduct several runs using different random seeds for each hyperparameter setting, and subsequently compute the mean performance. 
The hyperparameter setting that yields the highest mean performance is ultimately selected.

Each model's hyperparameter setting is a combination of several hyperparameters, which include the input dropout rate $dr_i$, the hidden dropout rate $dr_h$, the hidden dimensionality $d$, and the number of iterations $K$.
Enumerating all possible values of hyperparameters is impractical as the combination could result in thousands of distinct parameter settings.
To address this issue, we search for each hyperparameter only within a predetermined candidate range. 
This approach substantially decreases the number of potential parameter settings, thereby significantly reducing the time required for hyperparameter tuning.
We can roughly identify the candidate range of each parameter manually based on the performance under a set of randomly sampled parameter settings.
The candidate range for each model hyperparameter and the best-tuned model hyperparameters are provided in Appendix~\ref{app:params}.

\subsubsection{Evaluation Metrics}

For assessing the performance of the models, we employ evaluation metrics consistent with the settings of baseline models. 
The metrics are chosen as follows:
\begin{itemize}
\item For small datasets, specifically DBLP, IMDB, ACM, and Freebase, adherence is maintained to the HGB benchmark~\cite{lv2021we}. 
The metrics reported for these datasets are Macro-F1 and Micro-F1 scores, which evaluate the classification performance.
\item In the case of the OGBN-MAG dataset, the evaluation adheres to the methodology outlined in NARS~\cite{yu2020scalable} and SeHGNN~\cite{yang2023simple}, where the classification accuracy score is reported for this dataset.
\item For the OAG-Venue and OAG-L1-Field datasets, the evaluation follows the convention established by NARS~\cite{yu2020scalable}. 
The metrics reported for classification tasks on these datasets are Normalized Discounted Cumulative Gain (NDCG) and Mean Reciprocal Rank (MRR).
\end{itemize}

\begin{table*}[!tp]
  \centering
  \caption{
  Performance on Small and Large Datasets.
  }
  \vspace{-2mm}
\scalebox{0.65}{
\begin{tabular}{l c c c c c c c c c c c c c c}\hline
               &         & \multicolumn{2}{ c }{DBLP}      & \multicolumn{2}{c}{IMDB}        & \multicolumn{2}{c}{ACM}         & \multicolumn{2}{c}{Freebase} & OGBN-MAG & \multicolumn{2}{c}{OAG-Venue} & \multicolumn{2}{c}{OAG-L1-Field}\\\hline
               & E2E/Pre & Macro-F1       & Micro-F1       & Macro-F1       & Micro-F1       & Macro-F1       & Micro-F1       & Macro-F1       & Micro-F1       & accuracy   & ndcg      & mrr  & ndcg      & mrr\\\hline

GCN            & E2E     & 90.84$\pm$0.32 & 91.47$\pm$0.34 & 57.88$\pm$1.18 & 64.82$\pm$0.64 & 92.17$\pm$0.24 & 92.12$\pm$0.23 & 27.84$\pm$3.13 & 60.23$\pm$0.92 & OOM & OOM & OOM & OOM & OOM   \\
GAT            & E2E     & 93.83$\pm$0.27 & 93.39$\pm$0.30 & 58.94$\pm$1.35 & 64.86$\pm$0.43 & 92.26$\pm$0.94 & 92.19$\pm$0.93 & 40.74$\pm$2.58 & 65.26$\pm$0.80 & OOM & OOM & OOM & OOM & OOM   \\

GraphSAGE      & E2E     & 92.53$\pm$0.53 & 93.06$\pm$0.47 & 58.36$\pm$0.70 & 61.92$\pm$0.72 & 90.69$\pm$0.76 & 90.63$\pm$0.76 & 44.17$\pm$1.14 & 61.71$\pm$0.53 & 
                                                                                                                                                                   46.46$\pm$0.60 
                                                                                                                                                                              & 40.32$\pm$0.91
                                                                                                                                                                                           & 23.05$\pm$0.80 
                                                                                                                                                                                                  & 52.12$\pm$0.09 & 54.09$\pm$0.08 \\\hline

RGCN           & E2E     & 91.52$\pm$0.50 & 92.07$\pm$0.50 & 58.85$\pm$0.26 & 62.05$\pm$0.15 & 91.55$\pm$0.74 & 91.41$\pm$0.75 & 46.78$\pm$0.77 & 58.33$\pm$1.57 & 48.11$\pm$0.48 
                                                                                                                                                                              & 48.93$\pm$0.26
                                                                                                                                                                                          & 31.51$\pm$0.20
                                                                                                                                                                                                 & 85.91$\pm$0.10     
                                                                                                                                                                                                             & 84.92$\pm$0.23\\
HAN            & E2E     & 91.67$\pm$0.49 & 92.05$\pm$0.62 & 57.74$\pm$0.96 & 64.63$\pm$0.58 & 90.89$\pm$0.43 & 90.79$\pm$0.43 & 21.31$\pm$1.68 & 54.77$\pm$1.40 & OOM & OOM & OOM & OOM & OOM   \\
GTN            & E2E     & 93.52$\pm$0.55 & 93.97$\pm$0.54 & 60.47$\pm$0.98 & 65.14$\pm$0.45 & 91.31$\pm$0.70 & 91.20$\pm$0.71 & OOM            & OOM            & OOM & OOM & OOM & OOM & OOM   \\
RSHN           & E2E     & 93.34$\pm$0.58 & 93.81$\pm$0.55 & 59.85$\pm$3.21 & 64.22$\pm$1.03 & 90.50$\pm$1.51 & 90.32$\pm$1.54 & OOM            & OOM            & OOM & OOM & OOM & OOM & OOM   \\
HetGNN         & E2E     & 91.76$\pm$0.43 & 92.33$\pm$0.41 & 48.25$\pm$0.67 & 51.16$\pm$0.65 & 85.91$\pm$0.25 & 86.05$\pm$0.25 & OOM            & OOM            & OOM & OOM & OOM & OOM & OOM   \\
MAGNN          & E2E     & 93.28$\pm$0.51 & 93.76$\pm$0.45 & 56.49$\pm$3.20 & 64.67$\pm$1.67 & 90.88$\pm$0.64 & 90.77$\pm$0.65 & OOM            & OOM            & OOM & OOM & OOM & OOM & OOM  \\
HetSANN        & E2E     & 78.55$\pm$2.42 & 80.56$\pm$1.50 & 49.47$\pm$1.21 & 57.68$\pm$0.44 & 90.02$\pm$0.35 & 89.91$\pm$0.37 & OOM            & OOM            & OOM & OOM & OOM & OOM & OOM  \\
HGT            & E2E     & 93.01$\pm$0.23 & 93.49$\pm$0.25 & 63.00$\pm$1.19 & 67.20$\pm$0.57 & 91.12$\pm$0.76 & 91.00$\pm$0.76 & 29.28$\pm$2.52 & 60.51$\pm$1.16 & 46.78$\pm$0.42
                                                                                                                                                                               & 47.31$\pm$0.32  
                                                                                                                                                                                          & 29.82$\pm$0.33 
                                                                                                                                                                                            & 84.13$\pm$0.37 
                                                                                                                                                                                                & 82.16$\pm$0.89 \\

Simple-HGN     & E2E     & 94.01$\pm$0.24 & 94.46$\pm$0.22 & 63.53$\pm$1.36 & 67.36$\pm$0.57 & 93.42$\pm$0.44 & 93.35$\pm$0.45 & 47.72$\pm$1.48 & 66.29$\pm$0.45 & OOM & OOM & OOM & OOM & OOM  \\

HINormer       & E2E     & 94.57$\pm$0.23 & 94.94$\pm$0.21 & 64.65$\pm$0.53 & 67.83$\pm$0.34 & 92.23$\pm$0.40 & 92.19$\pm$0.40 & 49.94$\pm$2.12 & 66.47$\pm$0.47 & OOM & OOM & OOM & OOM & OOM \\\hline


NARS           & Pre     & 94.18$\pm$0.47 & 94.61$\pm$0.42  & 63.51$\pm$0.68 & 66.18$\pm$0.70 & 93.36$\pm$0.32 & 93.31$\pm$0.33  & 49.98$\pm$1.77 & 63.26$\pm$1.26 &50.66$\pm$0.22 & 52.28$\pm$0.17 
                                                                                                                                                                                           & 34.38$\pm$0.21 
                                                                                                                                                                                                 & 
                                                                                                                                                                                                
                                                                                                                                                                                                86.06$\pm$0.10      
                                                                                                                                                                                                             & 85.15$\pm$0.14\\


SeHGNN 
               & Pre     &  94.86$\pm$0.14 & 95.24$\pm$0.13 & 66.63$\pm$0.34 & 68.21$\pm$0.32 & 93.95$\pm$0.48 & 93.87$\pm$0.50 & 50.71$\pm$0.44 & 63.41$\pm$0.47 & 51.45$\pm$0.29 & 46.75$\pm$0.27 
                                                                                                                                                                                    & 29.11$\pm$0.25 
                                                                                                                                                                                        & 86.01$\pm$0.21 
                                                                                                                                                                                            & 84.95$\pm$0.20\\\hline
RpHGNN 
                & Pre     & \textbf{95.23$\pm$0.31} 
                                           & \textbf{95.55$\pm$0.29}
                                                            & \textbf{67.53$\pm$0.79}     
                                                                             & \textbf{69.77$\pm$0.66}   
                                                                                              &  \textbf{94.09$\pm$0.59} 
                                                                                                               & \textbf{94.04$\pm$0.59} 
                                                                                                                                & \textbf{54.02$\pm$0.88}    
                                                                                                                                                & \textbf{66.55$\pm$0.67} 
                                                                                                                                                            & \textbf{52.07$\pm$0.17} 
                                                                                                                                                                    & \textbf{53.31$\pm$0.40} 
                                                                                                                                                                                            & \textbf{35.46$\pm$0.47}
                                                                                                                  & \textbf{87.80$\pm$0.06}
                                                                                                                        & \textbf{86.79$\pm$0.18}\\\hline

\end{tabular}
}
\label{tab:performance}

  \vspace{-2mm}
  \end{table*}

\begin{table*}[!tp]
\vspace{-1mm}
  \centering
  \caption{
  Time Consumption (Pre-computation/Training/Overall Time Consumption (Seconds)) on Small and Large Datasets.
  }
\vspace{-2mm}
\scalebox{0.68}{
\begin{tabular}{l |c| c c c| c c c | c c c| c c c| c c c| c c c | c c c}\hline
           &          & \multicolumn{3}{c|}{DBLP}   & \multicolumn{3}{c|}{IMDB}   & \multicolumn{3}{c|}{ACM}  & \multicolumn{3}{c|}{Freebase} & \multicolumn{3}{c|}{OGBN-MAG}& \multicolumn{3}{c}{OAG-Venue} & \multicolumn{3}{c}{OAG-L1-Field}\\\hline
           & E2E/Pre & Pre   & Train & All   & Pre & Train & All  & Pre & Train & All  & Pre   & Train  & All  & Pre   & Train & All  & Pre   & Train & All   & Pre   & Train & All 
 \\\hline
GCN        & E2E     & 0     & 33    & 33    & 0   & 20    & 20   & 0   & 15    & 15   & 0     & 67     & 67   & OOM   & OOM   & OOM  & OOM   & OOM   & OOM   & OOM   & OOM   & OOM   \\
GAT        & E2E     & 0     & 33    & 33    & 0   & 21    & 21   & 0   & 26    & 26   & 0     & 63     & 63   & OOM   & OOM   & OOM  & OOM   & OOM   & OOM   & OOM   & OOM   & OOM   \\

GraphSAGE  & E2E     & 0     & 30    & 30    & 0   & 34    & 34   & 0   & 24    & 24   & 0     & 24     & 24   & 0     & 315  & 315 & 0     & 258   & 258   & 0     & 114   & 114  \\\hline
 
RGCN       & E2E     & 0     & 24    & 24    & 0   & 20    & 20   & 0   & 13    & 13   & 0     & 15     & 15   & 0     & 575   & 575  & 0     & 1538  & 1538  & 0     & 1535  & 1535  \\
HAN        & E2E     & 0     & 53    & 53    & 0   & 23    & 23   & 0   & 27    & 27   & 0     & 7518   & 7518 & OOM   & OOM   & OOM  & OOM   & OOM   & OOM   & OOM   & OOM   & OOM   \\
GTN        & E2E     & 0     & 14937 & 14937 & 0   & 5160  & 5160 & 0   & 1740  & 1740 & OOM   & OOM    & OOM  & OOM   & OOM   & OOM  & OOM   & OOM   & OOM   & OOM   & OOM   & OOM   \\
RSHN       & E2E     & 0     & 29    & 29    & 0   & 25    & 25   & 0   & 18    & 18   & OOM   & OOM    & OOM  & OOM   & OOM   & OOM  & OOM   & OOM   & OOM   & OOM   & OOM   & OOM   \\
HetGNN     & E2E     & 0     & 3101  & 3101  & 0   & 7260  & 7260 & 0   & 3208  & 3208 & OOM   & OOM    & OOM  & OOM   & OOM   & OOM  & OOM   & OOM   & OOM   & OOM   & OOM   & OOM   \\
MAGNN      & E2E     & 0     & 638   & 638   & 0   & 55    & 55   & 0   & 259   & 259  & OOM   & OOM    & OOM  & OOM   & OOM   & OOM  & OOM   & OOM   & OOM   & OOM   & OOM   & OOM   \\
HetSANN    & E2E     & 0     & 219   & 219   & 0   & 416   & 416  & 0   & 522   & 522  & OOM   & OOM    & OOM  & OOM   & OOM   & OOM  & OOM   & OOM   & OOM   & OOM   & OOM   & OOM   \\
HGT        & E2E     & 0     & 168   & 168   & 0   & 126   & 126  & 0   & 222   & 222  & 0     & 889    & 889  & 0     & 11704 & 11704& 0     & 12598 & 12598 & 0     & 16942 & 16942 \\

Simple-HGN & E2E     & 0     & 52    & 52    & 0   & 39    & 39   & 0   & 72    & 72   & 0     & 121    & 121  & OOM   & OOM   & OOM  & OOM   & OOM   & OOM   & OOM   & OOM   & OOM   \\
HINormer   & E2E     & 0     & 62    & 62    & 0   & 43    & 43   & 0   & 32    & 32   & 0     & 274    & 274  & OOM   & OOM   & OOM  & OOM   & OOM   & OOM   & OOM   & OOM   & OOM   \\\hline
NARS       & Pre     & 21    & 4     & 25    & 20  & 30    & 50   & 11  & 30    & 41   & 43    & 90     & 133  & 135   & 2220  & 2355 & 102   & 1044  & 1146  & 153   & 1425  & 1578  \\
SeHGNN     & Pre     & 21    & 21    & 42    & 18  & 103   & 121  & 9   & 77    & 86   & 56    & 559    & 615  & 19    & 2153  & 2172 & 10    & 2122  & 2132  & 17    & 4117  & 4134  \\\hline
RpHGNN     & Pre     & 32    & 4     & 36    & 25  & 14    & 39   & 12  & 17    & 29   & 53    & 13     & 66   & 102    & 1188  & 1290 & 100    & 723   & 823  & 106   & 1377  & 1483  \\\hline

\end{tabular}
\vspace{-2mm}
}
\label{tab:training_time}
\end{table*}

\subsection{Performance Analysis}

The performance of our approaches and baselines is shown in Table~\ref{tab:performance}.
Based on the results, we have the following observations:
\begin{itemize}
\item Heterogeneous GNNs demonstrate better overall performance compared to homogeneous GNNs.
This highlights the necessity of designing dedicated GNNs for heterogeneous graphs.
\item Pre-computation-based HGNNs show superior performance. 
The untrainable nature of pre-computation inhibits the adoption of learnable message passing. 
This encompasses techniques such as learnable message transformation, learnable input embeddings, and attention-based neighbor collection, all of which are extensively used by end-to-end baselines for extracting details from heterogeneous graphs.
Despite this, Table~\ref{tab:performance} shows that pre-computation-based HGNNs outperform end-to-end baselines in most cases.
This suggests that end-to-end trainability is not always essential.
Indeed, the untrainable message passing during pre-computation is sufficient for NARS, SeHGNN, and our approach to effectively capture enough details of heterogeneous graphs.
Note that the performance improvement of pre-computation-based HGNNs over end-to-end HGNNs is less significant on small datasets compared to large datasets.
This is understandable, as one of the main motivations for pre-computation-based HGNNs is to perform neighbor information collection across the entire graph, thereby alleviating the information loss caused by neighbor sampling operations required by end-to-end baselines on large datasets.
On small datasets, end-to-end baselines can be run directly on the entire graph without the need for neighbor sampling, thus minimizing information loss and diminishing the relative improvement offered by pre-computation-based HGNNs.
\item Neither the relation-wise style pre-computation-based HGNN (SeHGNN) nor the representation-wise style pre-computation-based HGNN (NARS) consistently outperforms the other across various datasets.
The two styles suffer from different types of information loss: SeHGNN struggles with information loss on large graphs due to its hop limitation, becoming inefficient when $K > 2$, while NARS encounters information loss due to the pre-merge of relations.
As shown in Table~\ref{tab:performance}, the relative performance of SeHGNN and NARS varies across datasets, highlighting the varying importance of different types of information loss in different graphs.
\item Our hybrid style shows superior performance over all baselines on all datasets.
In particular, by combining of the relation-wise style and representation-wise style, our approach is able to preserve the fine granularity of neighbor information and exploit relations with more hops.
Consequently, our hybrid style model outperforms both the relation-wise style model SeHGNN and the representation-wise style model NARS on all datasets.
\end{itemize}

\begin{figure}[!tp]
\vspace{-8mm}
\centering
\subfloat[
OGBN-MAG
]{
\includegraphics[width=1.6in]{fig_time_score_mag.pdf}
}
\subfloat[
OAG-Venue
]{
\includegraphics[width=1.6in]{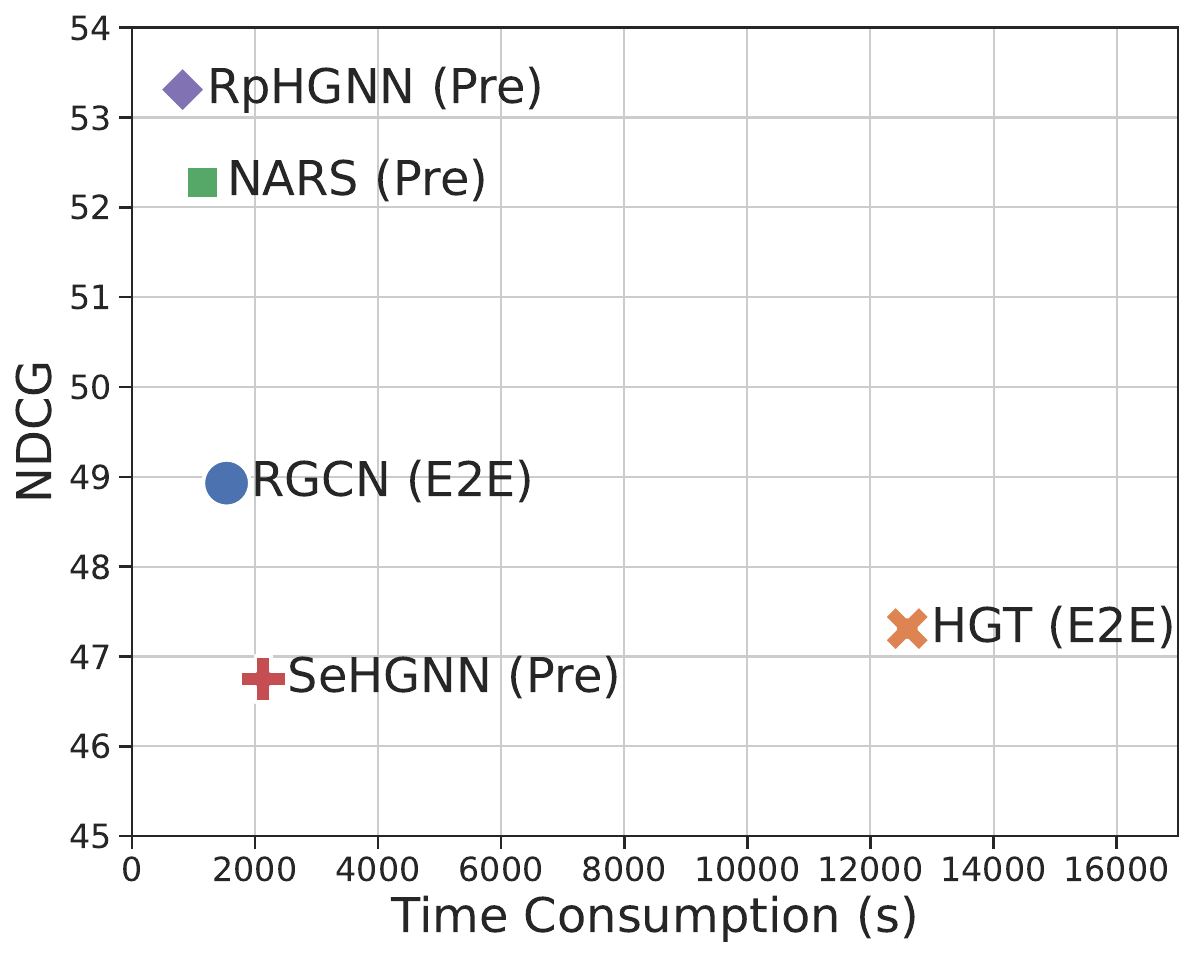}
}
\caption{
Performance and total time consumption of different methods on large datasets.
For end-to-end (E2E) methods, the time consumption refers to the training time.
For pre-computation-based (Pre) methods, the time consumption is the sum of pre-computation time and training time.
}
\label{fig:visualize_time_score} 
\vspace{-3mm}
\end{figure}

\begin{figure}[!tp]
\vspace{-6mm}
\centering
\subfloat[
OGBN-MAG
]{
\includegraphics[width=1.6in]{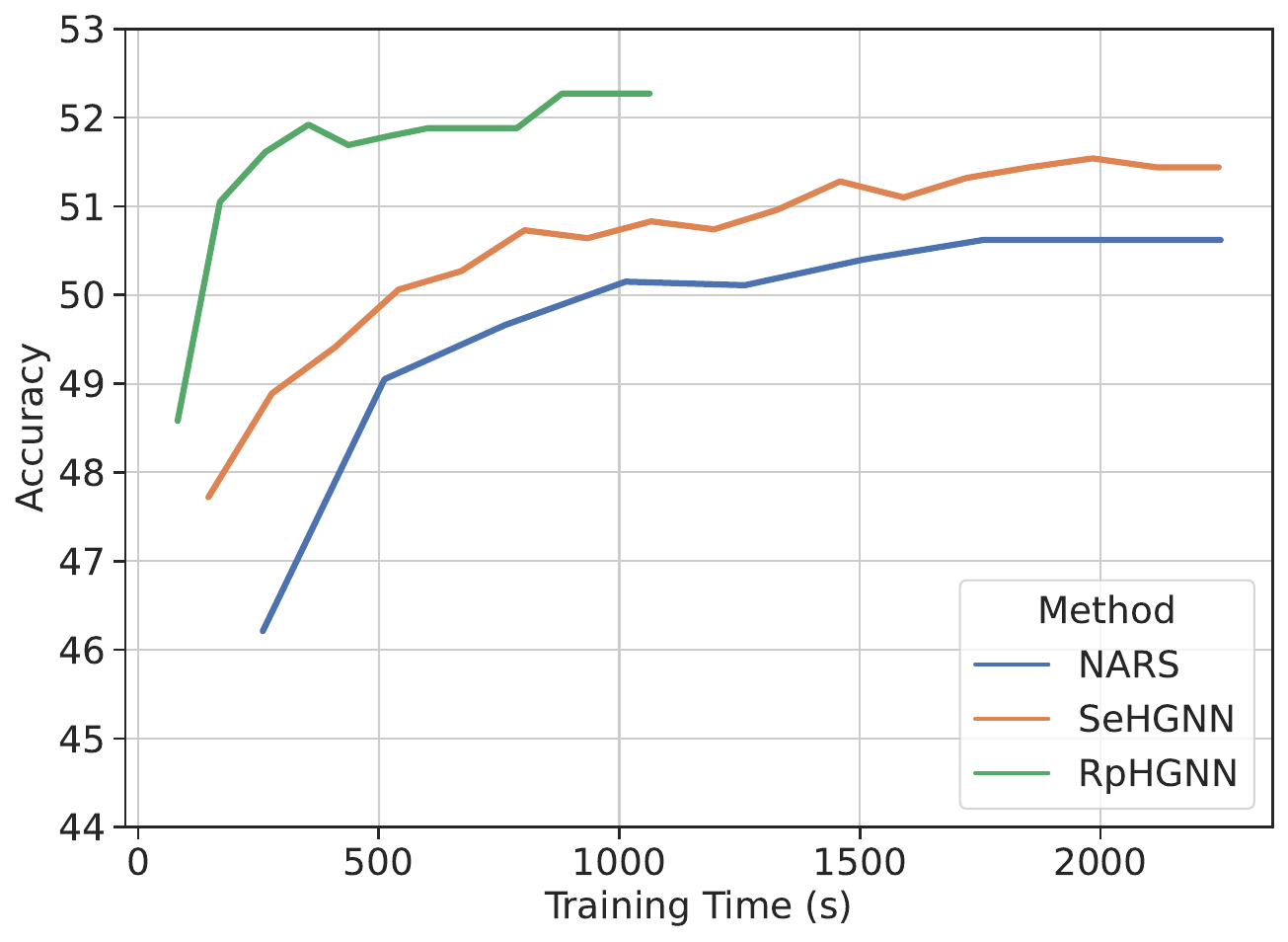}
}
\subfloat[
OAG-Venue
]{
\includegraphics[width=1.6in]{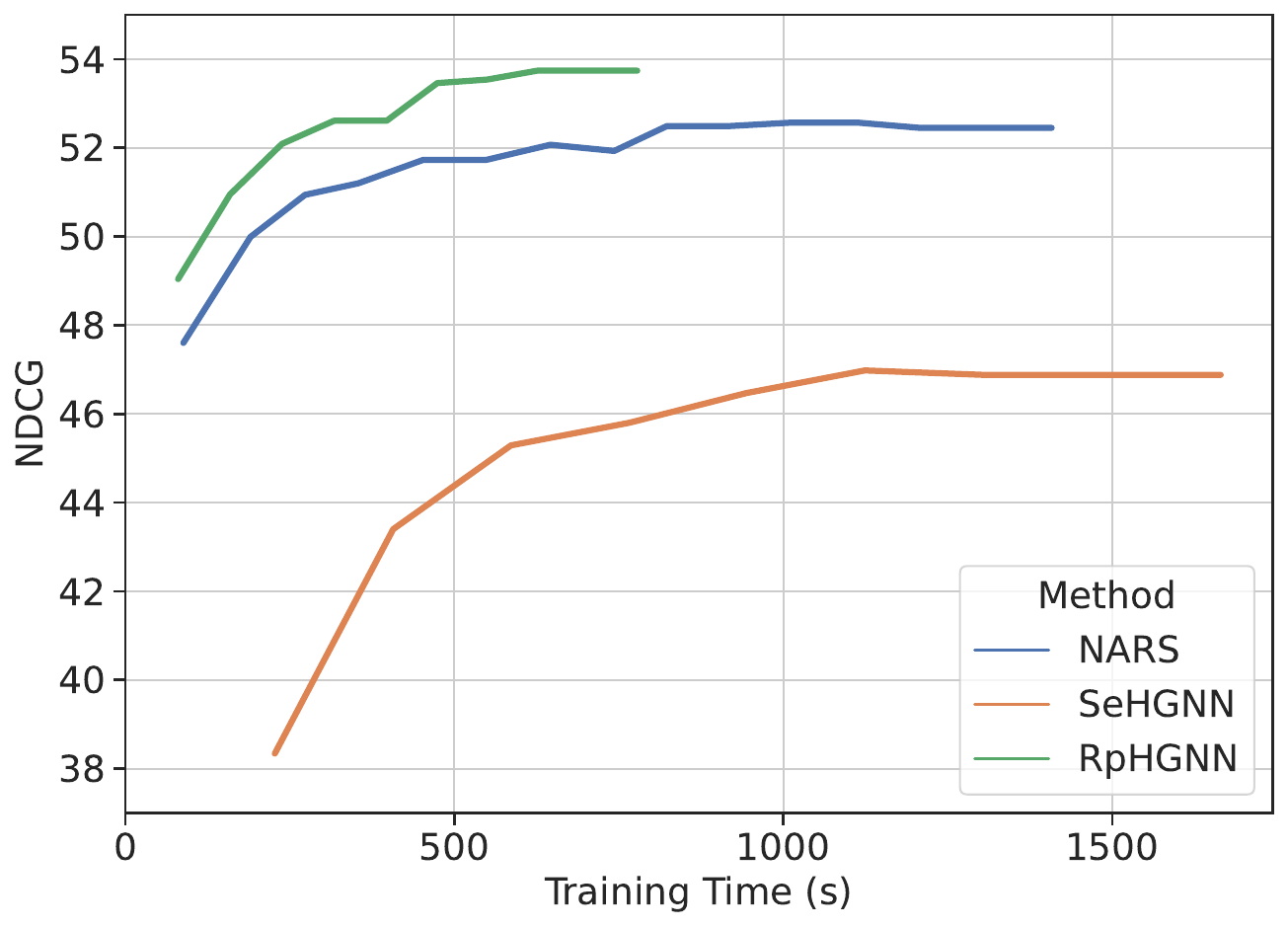}
}
\caption{
Test performance (recorded by the early stopping strategy) during training of different methods on large datasets.
}
\label{fig:visualize_training_curve} 
\vspace{-3mm}
\end{figure}

We compare the efficiency of our method and baselines in Table~\ref{tab:training_time}.
From the results, we observe:
\begin{itemize}
\item Pre-computation-based HGNNs exhibit scalability for large graphs, ensuring their efficient application to the large datasets OGBN-MAG, OAG-Venue, and OAG-L1-Field. 
Conversely, most end-to-end baselines encounter OOM issues when handling these datasets.
Furthermore, in comparison to the state-of-the-art end-to-end baseline, Simple-HGN (the state-of-the-art in terms of performance), our method is more efficient across most datasets, with the exception of the IMDB dataset, where both methods exhibit equal efficiency.
This highlights that by effectively extracting comprehensive details from heterogeneous graphs via pre-computation, our method can efficiently learn the semantics of vertices during the training stage.
\item Our pre-computation-based method shows superior efficiency over pre-computation-based baselines.
In terms of efficiency, our method outperforms NARS on six out of seven datasets and SeHGNN across all datasets.
Notably, the training time for our method is significantly less than that of NARS and SeHGNN on most datasets.
This indicates that the neighbor information gathered through our hybrid pre-computation method enhances training efficiency.
\item Our hybrid style HGNN exhibits superiority in overall effectiveness and efficiency. 
Considering all datasets, with performance surpassing the most effective baseline (SeHGNN) (as shown in Table~\ref{tab:performance}), RpHGNN also presents an improvement in overall speed compared to SeHGNN (as shown in Table~\ref{tab:training_time}).
To clearly show the overall effectiveness and efficiency of our method on large graphs, we visualize the test performance against overall time on the two large datasets for RGCN, HGT, NARS, SeHGNN, and our method in Fig.~\ref{fig:visualize_time_score}.
In addition, we also show the test performance during training in Fig.~\ref{fig:visualize_training_curve}.
The results show that our approach exhibits superiority in overall effectiveness and efficiency on large graphs.
\end{itemize}

\begin{table*}[!tp]
\centering
\caption{Ablation test of the even-odd propagation scheme.
We compare three propagation schemes: the local propagation scheme (with $2K$ iterations), the 2-hop propagation scheme (with $K$ iterations), and the even-odd propagation scheme (with $K$ iterations). 
All of them capture relations within $2K$ hops.
}

\scalebox{0.82}{
\begin{tabular}{l c c c c c c c c c c c c c}\hline
           & \multicolumn{2}{c}{DBLP}      & \multicolumn{2}{c}{IMDB}        & \multicolumn{2}{c}{ACM}         & \multicolumn{2}{c}{Freebase} & OGBN-MAG & \multicolumn{2}{c}{OAG-Venue} & \multicolumn{2}{c}{OAG-L1-Field}\\\hline
           & Macro-F1  & Micro-F1  & Macro-F1       & Micro-F1       & Macro-F1  & Micro-F1  & Macro-F1       & Micro-F1  & Accuracy   & NDCG      & MRR  & NDCG      & MRR \\\hline

RpHGNN-Local    & 95.02 &  95.35   & 66.24 & 68.96  & 93.58 & 93.51 & 53.32 & 66.35  & 51.03 & 52.84 & 34.92 & 87.79 & \textbf{86.85}\\\hline                
RpHGNN-2-Hop    & 95.22 &  95.54   & 66.79 & 69.08  & 94.07 & 94.02 & OOM   & OOM    & OOM   & OOM   & OOM   & OOM   & OOM\\\hline
RpHGNN-Even-Odd & \textbf{95.23} & \textbf{95.55} & \textbf{67.53} & \textbf{69.77} &  \textbf{94.09} & \textbf{94.04} & \textbf{54.02} & \textbf{66.55}   & \textbf{52.07} & \textbf{53.31} & \textbf{35.46} & \textbf{87.80} & 86.79 \\\hline
\end{tabular}
}
\label{tab:impact_even_odd}
\end{table*}

\begin{table}[!tp]
\centering
\caption{
Impact of Different Random Projection Strategies.
In Sparse Random Projection ($p^{sp}$), $p^{sp}$, as defined by Equation~\ref{eq:rand_proj_squash}, denotes the sparsity of the random projection weight matrix, and larger $p^{sp}$ indicates a more sparse random projection weight matrix.
}
\scalebox{0.9}{
\begin{tabular}{c c c c}
\hline
                                           & OGBN-MAG & \multicolumn{2}{c}{OAG-Venue}      \\\hline
Random Projection Strategies               & Accuracy & NDCG & MRR    \\\hline
Sparse Random Projection ($p^{sp} = 9/10$) & 52.10    & 53.35 & 35.51 \\\hline
Sparse Random Projection ($p^{sp} = 2/3$)  & 52.07    & 53.31 & 35.46 \\\hline
Sparse Random Projection ($p^{sp} = 1/2$)  & 52.19    & 53.43 & 35.56 \\\hline
Gaussian Random Projection                 & 52.04    & 53.28 & 35.44 \\\hline
\end{tabular}
}
\label{tab:impact_random_proj_strategy}
\end{table}

\subsection{Detailed Analysis}

In this section, we design different variants of our model to carry out detailed analyses on the effectiveness of the components of RpHGNN.

\subsubsection{Ablation test of Even-Odd Propagation Scheme}

We perform an ablation test of the even-odd propagation scheme to verify its effectiveness in Table~\ref{tab:impact_even_odd}. 
RpHGNN, which defaults to using the Even-Odd scheme, is aliased as RpHGNN-Even-Odd for clarity. 
RpHGNN-Local represents the variant with the local propagation scheme. 
RpHGNN-2-Hop utilizes a naive propagation scheme mentioned in Section~\ref{sec:even_odd_prop}, which simply extends local relations to all relations within 2 hops. 
We denote this propagation scheme as the 2-hop propagation scheme.

For a fair comparison, we adopt $2K$ iterations for RpHGNN-Local and $K$ iterations for RpHGNN-2-Hop and RpHGNN-Even-Odd ($K$ varies across different datasets). 
As a result, all of them can capture relations within $2K$ hops.

As indicated in Table~\ref{tab:impact_even_odd}, the performance of RpHGNN-Even-Odd surpasses that of RpHGNN-Local on six of the seven datasets. 
On the OAG-L1-Field dataset, the only exception, both methods exhibit comparable performance. 
This shows that by reducing the frequency of untrainable vertex representation update operations, the even-odd propagation scheme can effectively alleviate information loss to achieve superior performance.

Moreover, RpHGNN-Even-Odd achieves competitive performance with RpHGNN-2-Hop on three small datasets, but RpHGNN-2-Hop shows OOM on the other small datasets and all large datasets. 
This demonstrates that, by selecting only a subset of relations within 2 hops (even and odd relations), the even-odd propagation scheme shows efficiency advantages over the naive 2-hop propagation scheme.

\subsubsection{Impact of Random Projection Strategies for Random Projection Squashing}\label{sec:impact_rand_proj}

We compare the performance of our method when employing different random projection strategies for the Random Projection Squashing component.
These strategies include the sparse random projection strategy (with different sparsity $p^{sp}$) and the Gaussian random projection strategy.
Table~\ref{tab:impact_random_proj_strategy} reports the performance of each strategy.
Interestingly, the experiments reveal close performance across the different strategies, demonstrating the model’s insensitivity to the choice of random projection strategies.
This finding alleviates concerns about the inherent randomness in random projection operations used for untrainable vertex representation updates, which might have been suspected to cause performance fluctuations. 
The results indicate that this randomness does not significantly affect the semantics captured via pre-computation, thus maintaining consistent model performance.

\begin{figure}[!tp]
\vspace{-2mm}
\centering
\subfloat[
OGBN-MAG
]{
\label{fig:impact_norm_mag} 
\includegraphics[width=1.6in]{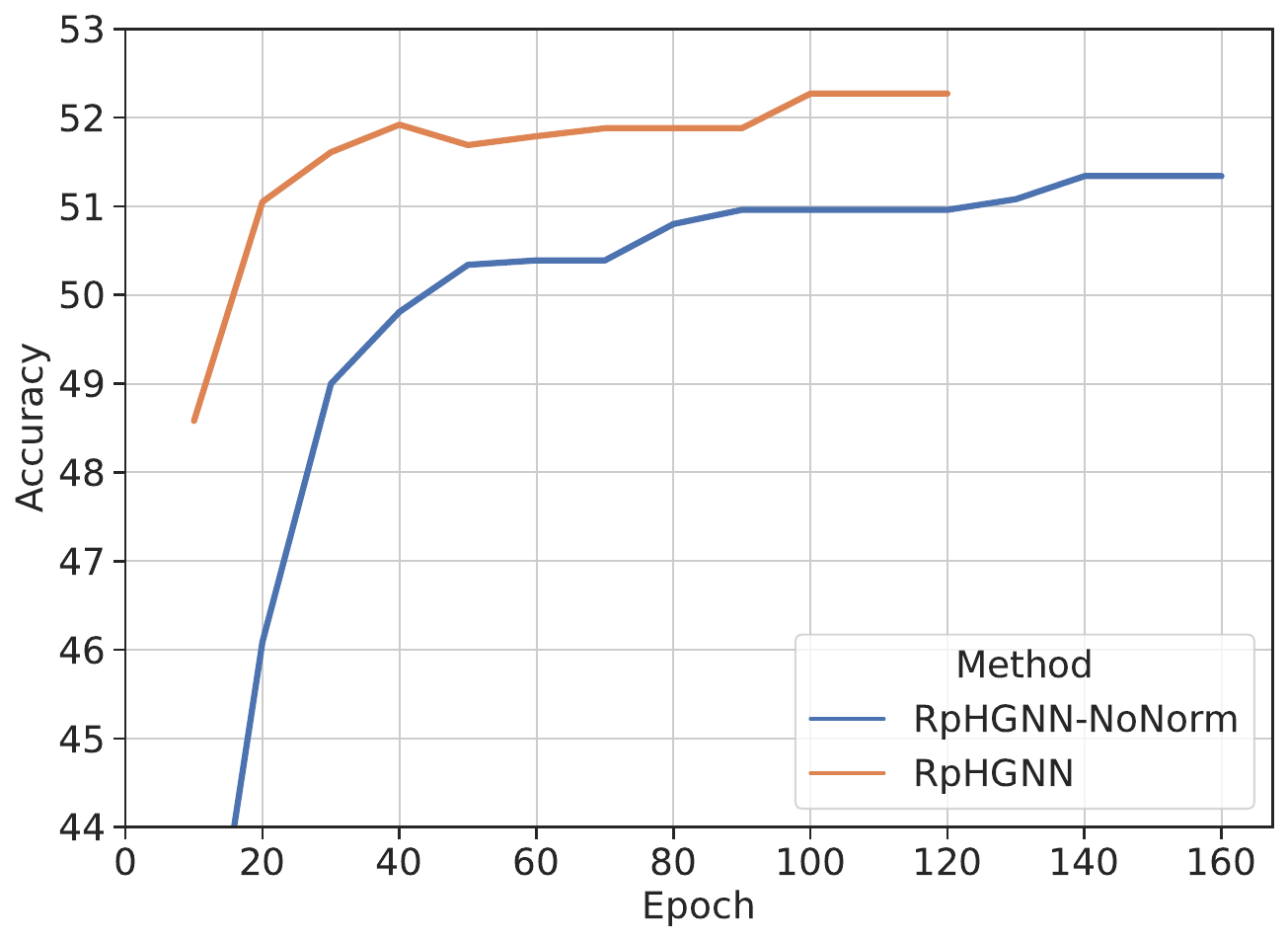}
}
\subfloat[
OAG-Venue
]{
\label{fig:impact_norm_oag_venue} 
\includegraphics[width=1.6in]{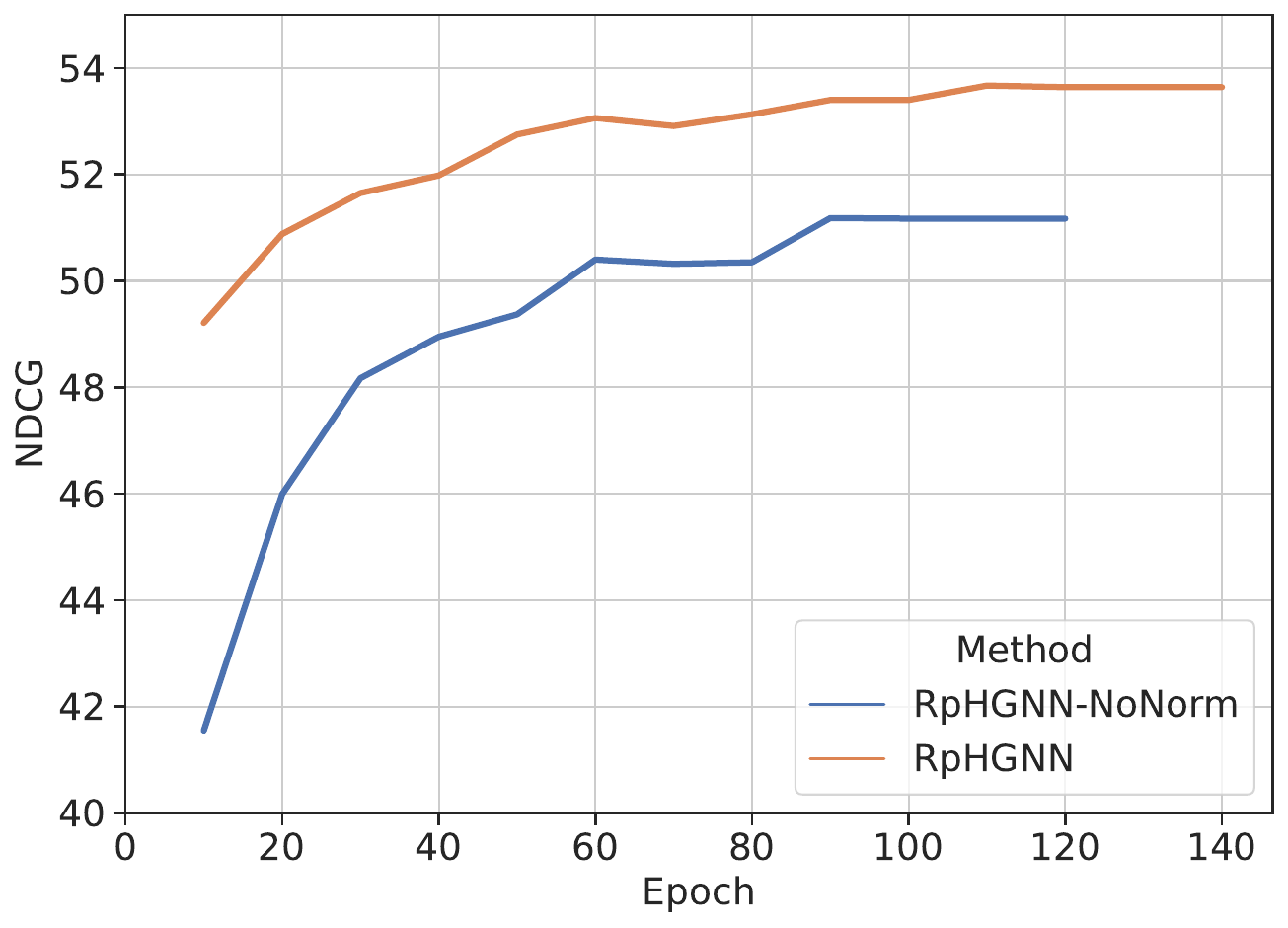}
}
\caption{
Impact of Normalization for Random Projection Squashing.
Test performance (recorded by the early stopping strategy) during training is reported.
}
\label{fig:impact_norm} 
\end{figure}

\subsubsection{Impact of Normalization for Random Projection Squashing}

We conduct an experiment to show the impact of the normalization operation in the Random Projection Squashing component, which aims to alleviate the information overwhelming problem of certain relations when squashing neighbor information collected via multiple relations.
We design a variant, RpHGNN-NoNorm, which removes the normalization in the Random Projection Squashing component.
Fig.~\ref{fig:impact_norm} compares the test performance during training for RpHGNN-NoNorm and RpHGNN.
The results show that, with the introduction of normalization, RpHGNN significantly outperforms RpHGNN-NoNorm, underscoring the efficacy of normalization in random projection squashing.

\subsubsection{Impact of Dimensionality of Random Embeddings for Featureless Vertices}

\begin{figure}[!tp]
\centering
\subfloat[
OGBN-MAG
]{
\label{fig:impact_dimension_mag} 
\includegraphics[width=1.5in]{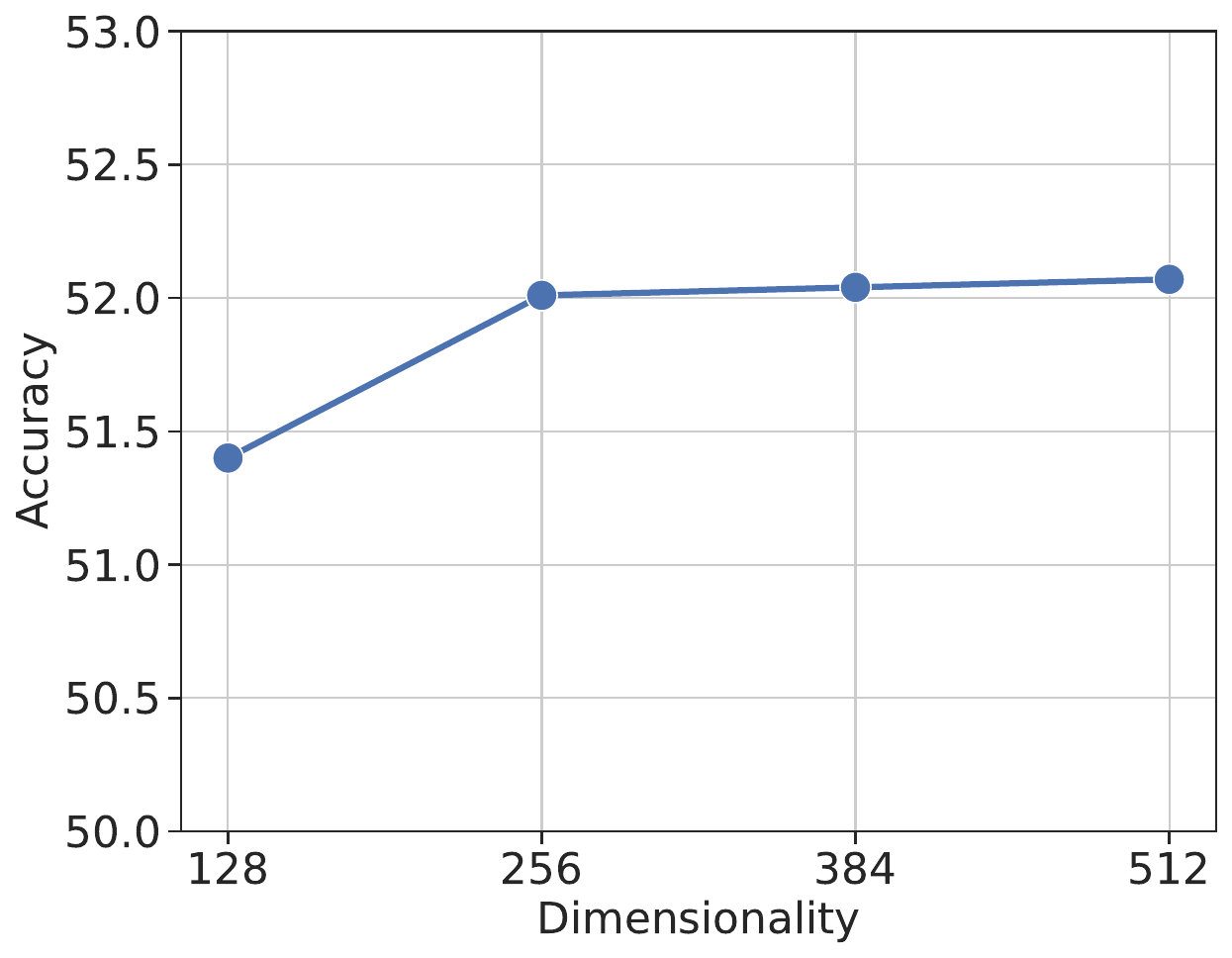}
}
\subfloat[
OAG-Venue
]{
\label{fig:impact_dimension_oag_venue} 
\includegraphics[width=1.5in]{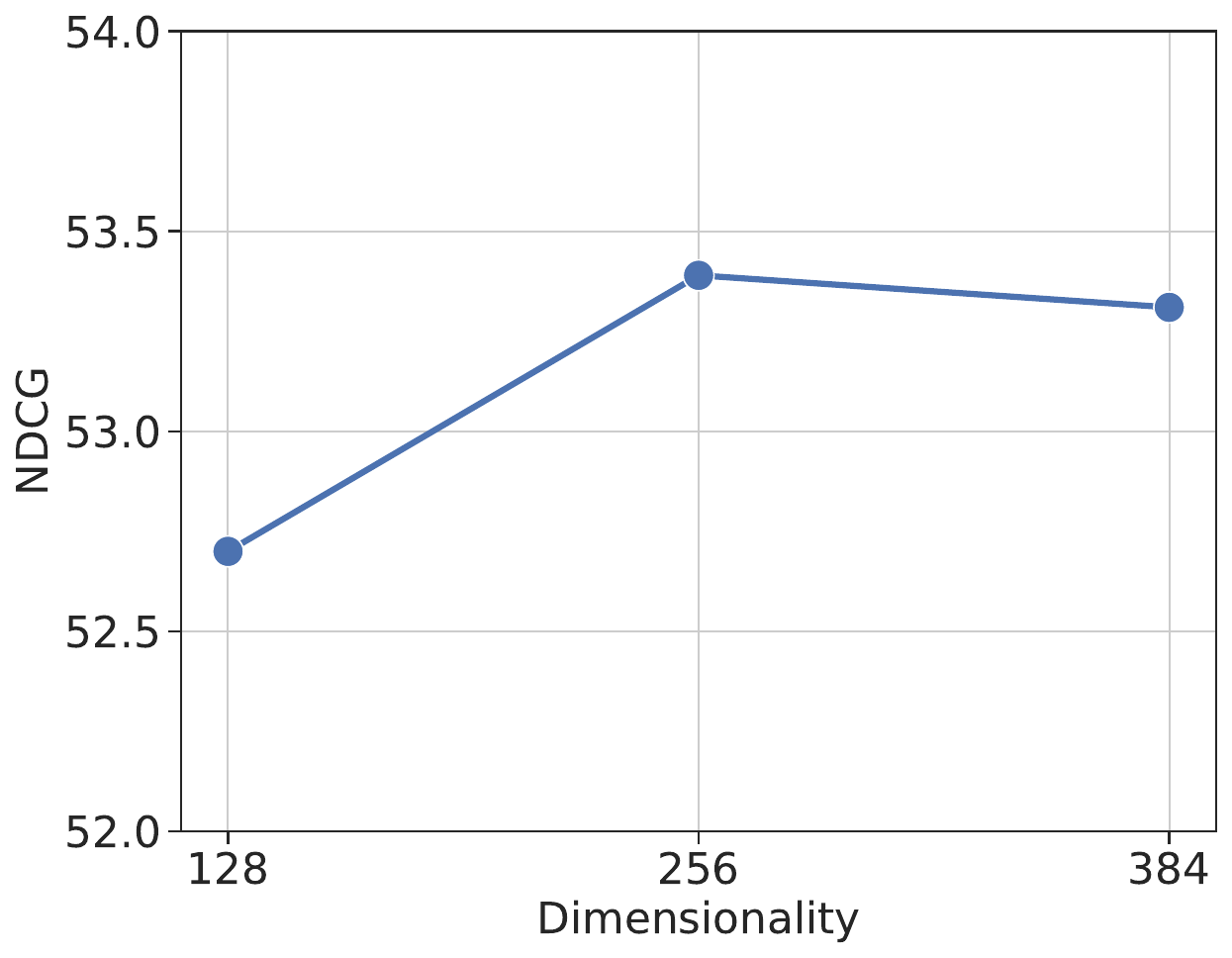}
}
\caption{
  Impact of Dimensionality of Random Embeddings for Featureless Vertices
}
\label{fig:impact_dimension} 
\end{figure}

As mentioned in the parameter setting section, for efficiency, we adopt random embeddings as input features for featureless vertices on large datasets.
For random embeddings, the most important parameter is their dimensionality $d$.
We visualize the impact of the dimensionality of random embeddings in Fig.~\ref{fig:impact_dimension}.
On the OGBN-MAG dataset, we vary $d$ from 128 to 512.
On the OAG-Venue dataset, since there are far more relations associated with the target vertices, we can only vary $d$ from 128 to 384.
The results show that for smaller values of $d$, an increase in dimensionality typically leads to significantly improved performance.
However, when $d$ is sufficiently large, the performance stabilizes, and no further obvious increase is observed.
This observation suggests that when sufficient dimensionality is provided for random embeddings, our model is able to effectively extract the semantics of heterogeneous graphs.

\subsubsection{Comparison with Top Method Combinations}\label{sec:leaderboard}

The OGBN-MAG benchmark dataset is associated with a public leaderboard\footnote{\url{https://ogb.stanford.edu/docs/leader_nodeprop/\#ogbn-mag}}.
In Table~\ref{tab:leaderboard}, we compare our method combination against top-performing approaches (method combinations) on the leaderboard, including NARS-GAMLP+LP+RLU+CR+Emb~\cite{zhang2021scr,zhang2022graph}, SeHGNN+LP+MS~\cite{yang2023simple}, and SeHGNN+LP+MS+Emb~\cite{yang2023simple}.
%
%
These methods integrate several general techniques such as Label Propagation (LP), Reliable Data Distillation (RLU), Consistency Regularization (CR), pre-trained Embeddings (Emb), and Multi-Stage Training (MS). 
By integrating CR and Emb, our approach RpHGNN+LP+CR+Emb outperforms other approaches, demonstrating its efficacy and compatibility with these techniques.

\section{Conclusion}

In this paper, we introduce Random Projection Heterogeneous Graph Neural Network (RpHGNN), a novel hybrid pre-computation-based HGNN, which achieves low information loss and efficiency by capitalizing on the advantages of two main styles of pre-computation-based HGNNs.
To achieve efficiency, the main framework of RpHGNN consists of propagate-then-update iterations, where we introduce a Random Projection Squashing step to ensure that complexity increases only linearly.
To achieve low information loss, we introduce a Relation-wise Neighbor Collection component with an Even-odd Propagation Scheme, which aims to collect information from neighbors in a finer-grained way. 
Experimental results illustrate that our approach achieves state-of-the-art results on seven small and large benchmark datasets while also being 230\% faster compared to the most effective baseline. 
This paper demonstrates that scalable HGNNs still hold potential for enhanced effectiveness through efficiency optimization. 
Efficiency optimizations, such as our random-projection-based dimensionality reduction strategy, enable HGNNs to access finer-grained information from additional hops of neighbors, thereby improving performance.
Additionally, we show that although untrainable operations during pre-computation, such as random projection, may introduce additional information loss, they can also be leveraged to design pre-computation-based HGNNs that are more effective and efficient.
As part of our future work, we aim to explore methods for further enhancing the performance and efficiency of RpHGNN, such as minimizing the size of collected neighbor information to reduce memory requirements during pre-computation and training. 
Additionally, we plan to investigate its applications in other tasks, such as link prediction.

\begin{table}[!tp]
\centering
\caption{
Performance on OGBN-MAG Leaderboard.
}
\scalebox{0.9}{
\begin{tabular}{l c c}\hline

                          & \multicolumn{2}{ c }{OGBN-MAG}                     \\\hline
                          & Validation Accuracy       & Test Accuracy                 \\\hline
NARS-GAMLP+LP+RLU+CR+Emb  &  57.34$\pm$0.35           & 56.31$\pm$0.21           \\
SeHGNN+LP+MS              &  58.70$\pm$0.08           & 56.71$\pm$0.14           \\
SeHGNN+LP+MS+Emb          &  59.17$\pm$0.09           & 57.19$\pm$0.12           \\
RpHGNN+LP+CR+Emb          & \textbf{59.73$\pm$0.08}   & \textbf{57.73$\pm$0.12}  \\\hline

\end{tabular}
}
\label{tab:leaderboard}
\end{table}

\section*{Acknowledgments}

This research is supported by the National Research Foundation, Singapore and Infocomm Media Development Authority under its Trust Tech Funding Initiative, the National Research Foundation, Singapore under its AI Singapore Programme (AISG Award No: AISG2-TC-2021-002), and the National Natural Science Foundation of China (Grant No: 62106262).

\bibliographystyle{IEEEtran}
\bibliography{citation}


\vfill

\clearpage
\newpage

\appendices

\markboth{Appendices to: Efficient Heterogeneous Graph Learning via Random Projection}%
{}

\appendices
\section{Best-tuned Model Hyperparameters}\label{app:params}

%
The candidate range of model hyperparameters are as follows:
\begin{itemize}
\item \textbf{DBLP}: $dr\_i \in \{0.7, 0.8\}$, $dr\_h \in \{0.7, 0.8\}$, $d \in \{256, 512\}$, $K=\{4, 5\}$.
\item \textbf{IMDB}: $dr\_i \in \{0.5, 0.6, 0.7, 0.8\}$, $dr\_h \in \{0.5, 0.6, 0.7, 0.8\}$, $d \in \{256, 512\}$, $K=\{2, 3\}$.
\item \textbf{ACM}: $dr\_i \in \{0.7, 0.8\}$, $dr\_h \in \{0.7, 0.8\}$, $d \in \{64, 128\}$, $K=\{5, 6\}$.
\item \textbf{Freebase}: $dr\_i \in \{0.7, 0.8\}$, $dr\_h \in \{0.7, 0.8\}$, $d \in \{256, 512\}$, $K=\{5, 6\}$.
\item \textbf{OGBN-MAG}: $dr\_i \in \{0.0, 0.1\}$, $dr\_h \in \{0.4, 0.5\}$, $d \in \{256, 512\}$, $K=\{4, 5\}$.
\item \textbf{OAG-Venue}: $dr\_i \in \{0.3, 0.5\}$, $dr\_h \in \{0.3, 0.5\}$, $d \in \{256, 512\}$, $K=\{2, 3\}$.
\item \textbf{OAG-L1-Field}: $dr\_i \in \{0.3, 0.5\}$, $dr\_h \in \{0.3, 0.5\}$, $d \in \{256, 512\}$, $K=\{2, 3\}$.
\end{itemize}

The best-tuned model hyperparameters are as follows:
\begin{itemize}
\item \textbf{DBLP}: $dr\_i=0.8$, $dr\_h=0.8$, $d=512$, $K=5$.
\item \textbf{IMDB}: $dr\_i=0.8$, $dr\_h=0.7$, $d=256$, $K=3$.
\item \textbf{ACM}: $dr\_i=0.8$, $dr\_h=0.7$, $d=64$, $K=6$.
\item \textbf{Freebase}: $dr\_i=0.7$, $dr\_h=0.7$, $d=256$, $K=6$.
\item \textbf{OGBN-MAG}: $dr\_i=0.0$, $dr\_h=0.5$, $d=512$, $K=5$.
\item \textbf{OAG-Venue}: $dr\_i=0.5$, $dr\_h=0.5$, $d=512$, $K=3$.
\item \textbf{OAG-L1-Field}: $dr\_i=0.3$, $dr\_h=0.5$, $d=512$, $K=3$.
\end{itemize}
Note that there is another model hyperparameter, $p^{sp}$, which denotes the sparsity of the random projection matrix.
As discussed in Section~\ref{sec:impact_rand_proj}, our model is insensitive to different random projection strategies, including variations in $p^{sp}$.
Therefore, we set $p^{sp}=2/3$ for all datasets.

\section{Even-Odd Propagation Scheme Affects Information Loss}\label{app:even-odd}

Given the example heterogeneous graph in Fig.~\ref{fig:framework}, we show how the even-odd propagation scheme compares to the local propagation scheme in terms of the information loss caused by untrainable vertex representation update.

\begin{figure}[!t]
\centering
\subfloat[
An example of how the local propagation scheme causes information loss due to untrainable vertex representation updates.
In the last column of the table, the number in parentheses () denotes how many untrainable vertex representation updates are performed when collecting neighbor information based on a relation.
For instance, (1) signifies that an untrainable vertex representation update occurs once when collecting neighbor information based on a relation.
Specifically, we use {\color{blue} (0, raw)} to highlight that the neighbor information is directly aggregated based on raw vertex features (without triggering any untrainable vertex representation update).
Given the target vertex $v_0$, we illustrate how its neighbor information is collected from $v_1$ and $v_2$ based on two specific relations f→p and p→a→p, respectively.
In the local propagation scheme, neighbor information can be collected via local relations without triggering untrainable vertex representation updates.
Thus, when collecting the neighbor $v_1$ for $v_0$ via the local relation f→p, the raw features of $v_1$ are utilized, and untrainable vertex representation updates, which can cause information loss, do not occur.
However, when collecting the neighbor $v_2$ of $v_0$ via p→a→p, the information of $v_2$ first traverses through p→a, then undergoes an untrainable vertex representation update (causing information loss), and finally flows via a→p to $v_0$.
\label{fig:prop_scheme_info_loss_local}
]{
\includegraphics[width=3.3in]{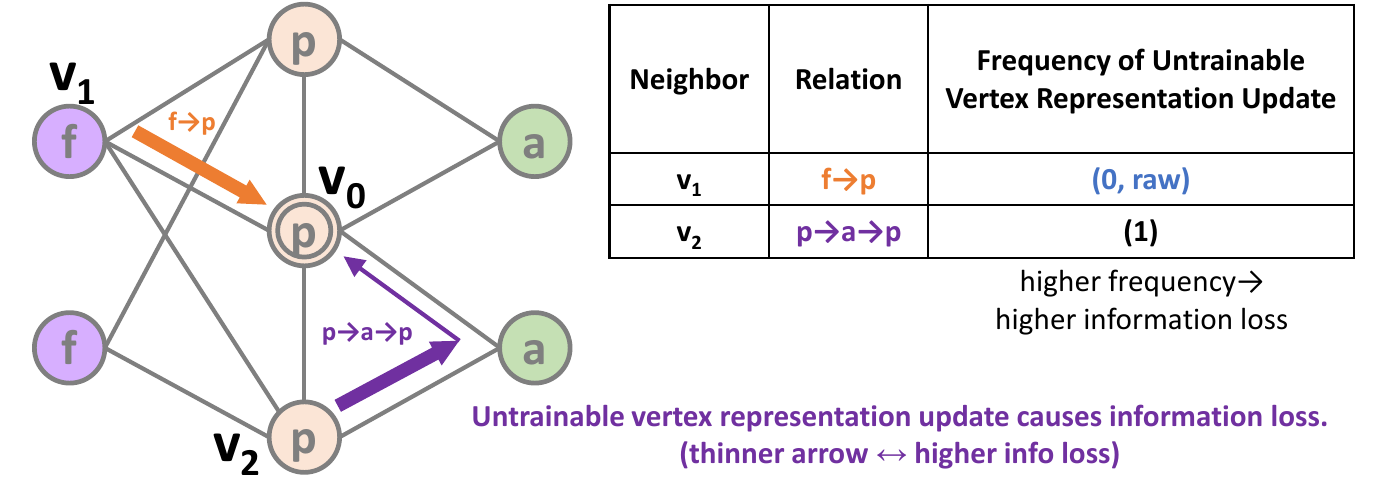}
}
\hspace{0.5mm}
\subfloat[
An example of how the even-odd propagation scheme alleviates information loss by reducing the frequency of untrainable vertex representation updates.
The detailed meaning of the number in parentheses () in the last column of the table is described in Fig.~\ref{fig:prop_scheme_info_loss_local}.
Given the target vertex $v_0$, we illustrate how its neighbor information is collected from $v_1$ and $v_2$ based on two specific relations f→p and p→a→p, respectively.
In the even-odd propagation scheme, neighbor information can be collected via either odd (local) or even relations without triggering untrainable vertex representation updates.
Thus, either when collecting the neighbor $v_1$ for $v_0$ via the odd (local) relation f→p, or when collecting the neighbor $v_2$ for $v_0$ via the even relation p→a→p, the raw vertex features can be utilized without any vertex representation update.
Compared to the local propagation scheme (Fig.~\ref{fig:prop_scheme_info_loss_local}), the even-odd propagation scheme may experience reduced information loss due to fewer untrainable vertex representation updates.\label{fig:prop_scheme_info_loss_even_odd}
]{
\includegraphics[width=3.3in]{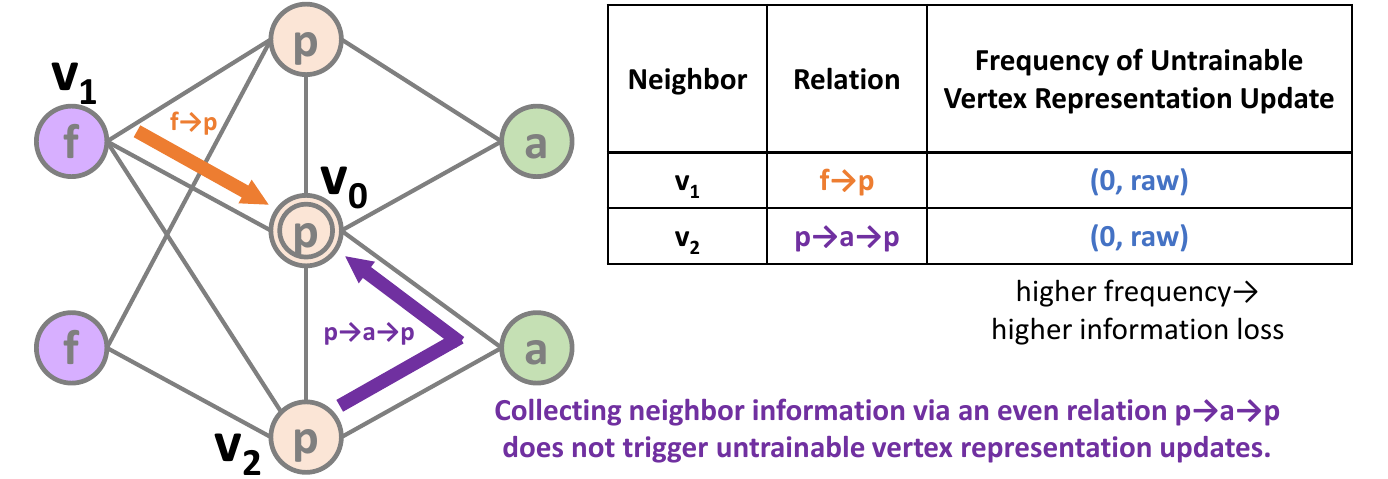}
}
\caption{
Information loss caused by untrainable vertex representation updates in local and even-odd propagation schemes.
}
\label{fig:prop_scheme_info_loss} 
\end{figure}

We first provide an intuitive example of the local propagation scheme and the even-odd propagation scheme in Fig.~\ref{fig:prop_scheme_info_loss_local} and Fig.~\ref{fig:prop_scheme_info_loss_even_odd}, respectively.
The details are provided in the captions of the figures.
Overall, compared to the local propagation scheme in Table~\ref{tab:pre_merge_local}, the even-odd propagation scheme shown in Table~\ref{tab:pre_merge_even_odd} can reduce information loss by adopting fewer untrainable vertex representation updates.

We provide another example to compare the local and even-odd propagation schemes more comprehensively.
For fairness, we compare a local propagation scheme with 4 iterations to an even-odd propagation scheme with 2 iterations, where both capture at most 4 hops of relations and collect 12 different vectors for each target vertex (each $\mathtt{paper}$ vertex).
As mentioned earlier, the untrainable vertex representation update in each propagate-then-update iteration can cause information loss in two ways: the pre-merge of relations and random projection operations.
We elaborate on these factors for the local and even-odd propagation schemes in Table~\ref{tab:pre_merge_local} and Table~\ref{tab:pre_merge_even_odd}, respectively.
The details are provided in the captions of the tables.
Overall, compared to the local propagation scheme in Table~\ref{tab:pre_merge_local}, the even-odd propagation scheme shown in Table~\ref{tab:pre_merge_even_odd} involes fewer untrainable vertex representation updates, reducing information loss caused by the pre-merge of relations and random projection operations. 
Furthermore, in Table~\ref{tab:impact_even_odd}, we show that in our framework, the even-odd propagation scheme (with $K/2$ iterations) can consistently outperform the local propagation scheme (with $K$ iterations), demonstrating that our framework can benefit from the low information loss brought by the even-odd propagation scheme.

\begin{table}[!tp]
\vspace{-40mm}
  \centering
  \caption{
  An example of neighbor information collected via the local propagation scheme for the heterogeneous graph in Fig.~\ref{fig:framework}.
  Each cell denotes the neighbor information collected via a relation (column) during a propagate-then-update iteration (row).
  In each cell, we list what relations are pre-merged and use the number in the parentheses () to denote how many untrainable vertex representation updates are performed when collecting neighbor information based on each relation.
  For clarity, we use $\mathtt{p}$, $\mathtt{f}$, $\mathtt{a}$, and $\mathtt{i}$ to denote vertex types $\mathtt{paper}$, $\mathtt{field}$, $\mathtt{author}$, and $\mathtt{institute}$, respectively.
  We highlight two cases with lower information loss as follows:
  {\color{red} Red} cells denote that the collected vector aggregates neighbor information from only one relation.
  {\color{blue} (0, raw)} cells denote that the neighbor information is directly aggregated based on raw vertex features (without triggering any untrainable vertex representation update).
  {\color{red} Red}+{\color{blue} (0,raw)} cells have the lowest information loss, which is preferred.
  }
    \label{tab:pre_merge_local}
    \scalebox{0.7}{
    
    \begin{tabular}{|p{18mm}|p{21mm}|p{21mm}|p{21mm}|}
    \hline
    \diagbox[width=22mm]{Iteration}{Relation} & p→p & f→p & a→p \\
    \hline
    1 & {\color{red} p→p} \newline {\color{blue} (0,raw)} 
                      & {\color{red} f→p} \newline {\color{blue} (0,raw)}
                                      & {\color{red} a→p} \newline {\color{blue} (0,raw)} \\\hline
    2 & p→p→p \newline 
        f→p→p \newline 
        a→p→p \newline
        (1)
                      & {\color{red} p→f→p} \newline
                        (1)
                                      & p→a→p \newline 
                                        i→a→p \newline
                                        (1) \\
    \hline
    3 & p→p→p→p \newline 
        f→p→p→p \newline
        a→p→p→p \newline
        p→f→p→p \newline
        p→a→p→p \newline
        i→a→p→p \newline
        (2)
                      & p→p→f→p \newline
                        f→p→f→p \newline
                        a→p→f→p \newline
                        (2)
                                      & p→p→a→p \newline
                                        f→p→a→p \newline
                                        a→p→a→p \newline
                                        a→i→a→p \newline 
                                        (2) \\
    \hline
    4 & p→p→p→p→p \newline
        f→p→p→p→p \newline
        a→p→p→p→p \newline
        p→f→p→p→p \newline
        p→a→p→p→p \newline
        i→a→p→p→p \newline
        p→p→f→p→p \newline
        f→p→f→p→p \newline
        a→p→f→p→p \newline
        p→p→a→p→p \newline
        f→p→a→p→p \newline
        a→p→a→p→p \newline
        a→i→a→p→p \newline
        (3)
                    & p→p→p→f→p \newline
                    f→p→p→f→p \newline
                    a→p→p→f→p \newline
                    p→f→p→f→p \newline
                    p→a→p→f→p \newline
                    i→a→p→f→p \newline
                    (3)
                                      & p→p→p→a→p \newline
                                        f→p→p→a→p \newline
                                        a→p→p→a→p \newline
                                        p→f→p→a→p \newline
                                        p→a→p→a→p \newline
                                        i→a→p→a→p \newline
                                        p→a→i→a→p \newline
                                        i→a→i→a→p \newline
                                        (3) \\
    \hline
    \end{tabular}
    }
    \end{table}

  \begin{table}[!tp]
  \vspace{-90mm}
  \centering
  \caption{
  An example of neighbor information collected via our even-odd propagation scheme.  
  The meaning of each cell, {\color{red} red}, and {\color{blue} blue} are described in Table~\ref{tab:pre_merge_local}.
  Compared to the local propagation scheme in Table~\ref{tab:pre_merge_local}, the even-odd propagation scheme involves less untrainable vertex representation updates, reducing information loss caused by the pre-merge of relations and random projection operations.
  Compared to the local propagation scheme, our even-odd propagation scheme results in more cells with minimal information loss (satisfying {\color{red} red }+{\color{blue} (0,raw)}), which is preferred.
  }
  \label{tab:pre_merge_even_odd}
  \scalebox{0.64}{
      
      \begin{tabular}{|p{18mm}|p{13mm}|p{13mm}|p{13mm}|p{17mm}|p{17mm}|p{17mm}|}
      \hline
      \diagbox[width=22mm]{Iteration}{Relation} & p→p & f→p & a→p & p→p→p & p→f→p  & p→a→p \\
      \hline
      1 & {\color{red} p→p} \newline {\color{blue} (0,raw)} 
          & {\color{red} f→p} \newline {\color{blue} (0,raw)} 
              & {\color{red} a→p} \newline {\color{blue} (0,raw)} 
                  & {\color{red} p→a→p} \newline {\color{blue} (0,raw)} 
                      & {\color{red} p→f→p} \newline {\color{blue} (0,raw)}                  
                          & {\color{red} p→p→p} \newline {\color{blue} (0,raw)} \\\hline
      2 &
      p→p→p \newline
      f→p→p \newline
      a→p→p \newline
      p→a→p→p \newline
      p→f→p→p \newline
      p→p→p→p \newline 
      (1,raw)
          & p→f→p \newline
            f→p→f→p \newline 
            (1,raw)
            & p→a→p \newline
              i→a→p \newline
              a→p→a→p \newline
              a→i→a→p \newline 
              (1,raw)
              & p→p→p→p \newline
                f→p→p→p \newline
                a→p→p→p \newline
                p→a→p→p→p \newline
                p→f→p→p→p \newline
                p→p→p→p→p \newline 
                (1,raw)
                & p→p→f→p \newline
                  f→p→f→p \newline
                  a→p→f→p \newline
                  p→a→p→f→p \newline
                  p→f→p→f→p \newline
                  p→p→p→f→p \newline 
                  (1,raw)
                  & p→p→a→p \newline
                    f→p→a→p \newline
                    a→p→a→p \newline
                    p→a→p→a→p \newline
                    p→f→p→a→p \newline
                    p→p→p→a→p \newline 
                    (1,raw) \\
      \hline
      \end{tabular}
      }
      \vspace{-2mm}
      \end{table}

\end{document}